\pdfoutput=1
\documentclass{article}

\PassOptionsToPackage{numbers,compress}{natbib}  
\usepackage[eandd, preprint]{neurips_2026}

\usepackage[utf8]{inputenc}
\usepackage[T1]{fontenc}
\usepackage{amsmath,amssymb}
\usepackage{graphicx}
\usepackage{booktabs}
\usepackage{hyperref}
\usepackage{microtype}
\usepackage{xcolor}
\usepackage{multirow}
\usepackage{caption}
\usepackage{enumitem}
\usepackage{titlesec}
\usepackage{longtable}
\usepackage{listings}
\usepackage{verbatim}

\usepackage{todonotes}

\titlespacing*{\section}{0pt}{5pt plus 1pt minus 1pt}{1pt plus 1pt minus 0.5pt}
\titlespacing*{\subsection}{0pt}{3pt plus 1pt minus 1pt}{1pt plus 1pt minus 0.5pt}
\titlespacing*{\paragraph}{0pt}{2pt plus 1pt minus 1pt}{0.3em}

\setlist[itemize]{itemsep=0pt, parsep=0pt, topsep=0pt}
\setlist[enumerate]{itemsep=0pt, parsep=0pt, topsep=0pt}

\usepackage{xr}
\makeatletter
\renewcommand{\@maketitle}{%
  \vbox{%
    \hsize\textwidth
    \linewidth\hsize
    \vskip 0.05in
    \@toptitlebar
    \centering
    {\LARGE\bf \@title\par}
    \@bottomtitlebar
    \def\And{%
      \end{tabular}\hfil\linebreak[0]\hfil%
      \begin{tabular}[t]{c}\bf\rule{\z@}{14\p@}\ignorespaces%
    }
    \def\AND{%
      \end{tabular}\hfil\linebreak[4]\hfil%
      \begin{tabular}[t]{c}\bf\rule{\z@}{14\p@}\ignorespaces%
    }
    \begin{tabular}[t]{c}\bf\rule{\z@}{14\p@}\@author\end{tabular}%
    \vskip 0.10in \@minus 0.05in
  }
}
\renewenvironment{abstract}%
{%
  \vskip 0.02in%
  \centerline{\large\bf Abstract}%
  \vspace{0.2ex}%
  \begin{quote}%
}{%
  \par%
  \end{quote}%
  \vskip 0.5ex%
}
\makeatother

\title{ChronoMedKG: A Temporally-Grounded Biomedical Knowledge Graph and Benchmark for Clinical Reasoning}

\author{%
  Md Shamim Ahmed\thanks{Corresponding author: \texttt{shamim@imada.sdu.dk}} \\
  University of Southern Denmark \\
  \And
  Farzaneh Firoozbakht \\
  University of Hamburg \\
  \And
  Lukas Galke Poech \\
  University of Southern Denmark \\
  \AND
  Jan Baumbach \\
  University of Hamburg \\
  \And
  Richard R\"{o}ttger \\
  University of Southern Denmark \\
}

\begin{document}

\maketitle

\begin{abstract}
Biomedical knowledge graphs (KGs) treat disease associations as static facts, but temporal information is crucial for clinical reasoning, e.g., a symptom diagnostic of one disease at age 3 may imply a different disease at age 13. Existing KGs such as PrimeKG, Hetionet, and iKraph do not encode \emph{when} a finding becomes clinically relevant over the course of a disease. This limits their usefulness for longitudinal clinical reasoning and retrieval augmentation.
We introduce \textbf{ChronoMedKG}, a temporal biomedical knowledge graph that contains 460{,}497 evidence-linked triples (filtered from 13M raw extractions) covering 13{,}431 diseases. Each association is tied to temporal components like onset window or progression stage, which are backed by PMID-traceable evidence and a multi-signal credibility score. The graph is constructed through a disease-autonomous multi-agent pipeline in which multiple frontier LLMs independently extract knowledge from PubMed and PMC literature. Only those relations are kept that are supported by multi-model consensus, survive credibility filtering, as well as ontology alignment.
ChronoMedKG scored 92.7\% agreement against Orphadata and adds temporal grounding for 6{,}250 diseases absent from HPOA, Orphadata, and Phenopackets, including 1{,}657 Orphanet-coded rare diseases. We further introduce \textbf{ChronoTQA}, a benchmark of 3{,}341 questions across eight task types (six temporal plus two static controls), with a 12-question supplementary probe. Frontier LLMs lose roughly 30 points moving from static to temporal questions; ChronoMedKG retrieval rescues 47--65\% of their long-tail failures, against 17--29\% for HPOA-RAG. As such, ChronoMedKG provides a crucial temporal axis for retrieval-augmented clinical systems that was previously absent. 
\end{abstract}

\begin{center}
\footnotesize
\textbf{Dataset:}~\url{https://doi.org/10.5281/zenodo.19697542}\hspace{1.5em}\textbf{Code:}~\url{https://gitlab.sdu.dk/screen4care/chronomedkg}
\end{center}

\section{Introduction}
\label{sec:intro}
Biomedical knowledge graphs (KGs) are widely used in applications such as drug repurposing and computational phenotyping, and are increasingly explored for supporting AI-driven clinical reasoning \citep{chandak2023primekg,himmelstein2017hetionet,su2025ikraph}. For example, PrimeKG \citep{chandak2023primekg} integrates 20 curated databases into 4.05M edges across 17{,}080 diseases; iKraph \citep{su2025ikraph}, Hetionet \citep{himmelstein2017hetionet}, and KARMA \citep{karma2025} take alternative integration approaches.
However, these KGs lack temporal dimension of clinical knowledge. For example, associations such as "DMD is associated with cardiomyopathy" do not convey when manifestations occur or how evidence evolves over time. Static edges do not capture such temporal progression and do not distinguish whether the association comes from a 1995 case series or a 2023 cohort study. Yet clinical decisions depend on this temporal context: when to start screening, which diagnosis to consider at a given age, and whether intervention is still early enough to matter.

Existing resources provide limited support for temporal clinical information. The most temporally-aware structured resource, HPOA \citep{hpoa}, provides coarse onset categories (e.g., ``childhood'') for 1{,}429 diseases, while Orphadata covers onset for 5{,}796 rare diseases, both at the disease level rather than phenotype level. In addition, Phenopackets \citep{phenopackets} provides patient-level case data for 518 diseases. Even when combined, these sources cover at most 7{,}743 diseases out of PrimeKG's 17{,}080 and do not provide per-phenotype temporal annotations at scale. No biomedical QA benchmark (e.g., MedQA, PubMedQA, BioASQ, MMLU-Medical \citep{medqa,pubmedqa,bioasq,hendrycks2021mmlu}) evaluates temporal clinical reasoning either. The temporal dimension of disease progression is therefore both unstructured and untested.

\noindent\textbf{Contributions.}
\begin{enumerate}[leftmargin=*,itemsep=1pt]
    \item \textbf{ChronoMedKG}: a temporally-grounded biomedical KG of 460{,}497 validated consensus triples derived from 13{,}431 diseases, built by a disease-autonomous four-agent pipeline with multi-LLM consensus (Section~\ref{sec:construction}). Each triple includes onset ages, progression stages, clinical milestones, and six-signal evidence grading with PMID provenance. We validate against Orphadata at \textbf{92.7\%} effective accuracy on the 2{,}563-disease overlap (Section~\ref{sec:validation}, Table~\ref{tab:validation}) and via an independent three-LLM judge-panel audit at \textbf{87.9\%} verified accuracy on a 100-disease novel-coverage sample (Section~\ref{sec:novelty}, Table~\ref{tab:novelty}); neither figure is a claim about the full 13{,}431-disease resource, and clinician validation at scale across the 6{,}250 novel-coverage diseases is the remaining step (Limitation~(ii), Section~\ref{sec:discussion}).
    \item \textbf{ChronoTQA} (Section~\ref{sec:benchmark}). To the best of our knowledge, ChronoTQA is the first temporal biomedical QA benchmark, with 3{,}329 questions across eight reported task types plus a 12-question supplementary diagnostic set; Tier~1 questions are grounded in Orphadata, HPOA, and GA4GH Phenopackets, and Tier~2 questions trace to PMID-verified ChronoMedKG triples.
    \item \textbf{Three claims that static KGs cannot support} (Section~\ref{sec:novel}). A coverage-gap analysis shows 6{,}250 diseases with onset data absent from every curated resource, of which 1{,}657 are Orphanet-coded rare diseases. Retrieval against ChronoMedKG rescues 47--65\% of long-tail onset queries that three frontier LLMs answer incorrectly without retrieval. A frontier-LLM baseline records a $+$30~pp gap between static- and temporal-question accuracy.
\end{enumerate}

\noindent\textbf{Scope of claims.} ChronoMedKG is built from peer-reviewed literature, not patient records, and any clinical use requires clinician oversight and regulatory evaluation beyond the scope of this work. The validation harness, judge-panel code, and error-taxonomy classifier are released so the community can audit and extend these numbers directly.

\begin{figure}[t]
    \centering
    \includegraphics[width=\linewidth]{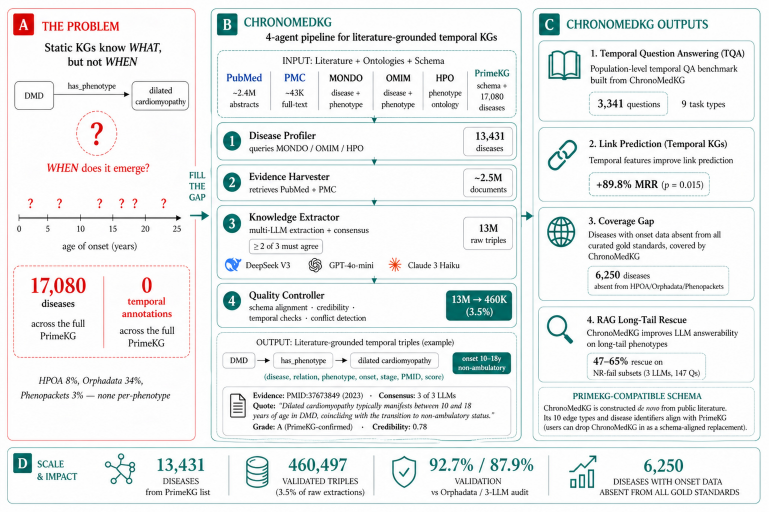}
    \caption{\textbf{ChronoMedKG: from \emph{what} is associated to \emph{when} it emerges.} \textbf{(A)}~Across 17{,}080 PrimeKG diseases, zero edges carry temporal annotations. \textbf{(B)}~4-agent pipeline ingests $\sim$2.5M public documents and produces literature-grounded temporal triples with PMID provenance, multi-LLM consensus, and six-signal credibility scoring. \textbf{(C)}~Downstream outputs unique to ChronoMedKG. \textbf{(D)}~Top-line numbers. Constructed \emph{de novo} from public literature; no patient-level records.}
    \label{fig:pipeline}
\end{figure}

\section{Related Work}
\label{sec:related}

\paragraph{Biomedical knowledge graphs.}
Table~\ref{tab:comparison} summarizes existing biomedical KGs. PrimeKG \citep{chandak2023primekg} integrates 20 curated databases into a 4.05M-edge graph across 17{,}080 diseases. iKraph \citep{su2025ikraph} extracts over 30M relations from PubMed and reports strong performance in drug repurposing; however, it provides only document-level timestamps (publication dates) rather than clinically meaningful temporal information. Hetionet \citep{himmelstein2017hetionet} covers 136 diseases with 2.25M edges. KARMA \citep{karma2025} (NeurIPS 2025 Spotlight) introduces a multi-agent collaborative pipeline to extract 150K triples from 1{,}200 articles with multi-agent conflict resolution; it validates at 83.1\% LLM-verified accuracy but does not incorporate temporal metadata or provide a downstream temporal evaluation. \textbf{None of these resources captures \emph{when} clinical associations occur during disease progression.}

\paragraph{Concurrent agentic-extraction work.}
Two recent systems extend biomedical KG construction along axes orthogonal to ChronoMedKG. MedKGent \citep{medkgent2025} (Aug 2025) processes 10M PubMed abstracts as a 1975--2023 daily time series, producing a 2.97M-triple medical KG with self-consistency confidence scoring. Its temporal axis is the \emph{publication date} of source abstracts: when knowledge appeared in the literature. ChronoMedKG instead grounds facts in patient age and disease stage. AutoBioKG \citep{autobiokg2026} (bioRxiv Jan 2026) takes a different angle: each edge carries composite contextual conditions, so a relation can flip under different physiological states (Ca\textsuperscript{2+}-gated activation/inhibition is the running example). ChronoMedKG addresses an analogous expressivity gap by encoding context as structured ontology fields (\texttt{onset\_age\_min}/\texttt{onset\_age\_max}, named \texttt{progression\_stage}, named \texttt{milestone}), not free-text strings. Publication-time, contextual-state, and clinical-time are three independent axes that could be unified in future work.

\begin{table}[t]
\centering
\small
\caption{ChronoMedKG vs.\ existing biomedical KGs. Only ChronoMedKG provides per-phenotype temporal metadata, edge-level PMID + verbatim quotes, and a dedicated temporal QA benchmark. \textsuperscript{*}MedKGent disease count estimated from a 156K total-entity release without node-type breakdown.}
\label{tab:comparison}
\begin{tabular}{lcccccc}
\toprule
\textbf{Resource} & \textbf{Diseases} & \textbf{Edges} & \textbf{PMID per} & \textbf{Evidence} & \textbf{Temporal} & \textbf{Temporal} \\
 & & & \textbf{edge} & \textbf{quote} & \textbf{profiles} & \textbf{QA benchmark} \\
\midrule
iKraph & $\sim$48K & 30.8M & Partial & No & No & No \\
PrimeKG & 17{,}080 & 4.05M & No & No & No & No \\
MedKGent\textsuperscript{*} & $\sim$14K & 2.97M & Partial & No & Pub-date only & No \\
Hetionet & $\sim$4{,}500 & 2.25M & No & No & No & No \\
HPOA & $\sim$8{,}600 & 283K & Yes (ref col) & No & Coarse & No \\
KARMA & $\sim$200 & $\sim$150K & Partial & No & No & No \\
\midrule
\textbf{ChronoMedKG} & \textbf{13{,}431} & \textbf{460K} & \textbf{100\%} & \textbf{99.9\%} & \textbf{Per-phenotype} & \textbf{ChronoTQA (3{,}341 Qs)} \\
\bottomrule
\end{tabular}
\end{table}

\paragraph{Temporal knowledge graphs.}
Prior work on temporal KGs \citep{trivedi2017knowevolve,dasgupta2018hyte} models time-stamped facts (e.g., ``Company X acquired Company Y in 2019'') where the temporal dimension is the calendar date of an event. Clinical temporal grounding differs in two ways: the temporal dimension is \emph{patient age} or \emph{disease stage} at which a phenotype manifests, and the underlying source is population-level literature rather than per-event records. T-Phenotype \citep{t_phenotype2023} discovers temporal phenotypes from patient-level EHR data using deep clustering, which is complementary to our literature-derived approach. To our knowledge, no prior work has built a population-level temporally-grounded biomedical KG from literature at scale.

\paragraph{Biomedical QA and evaluation benchmarks.}
MedQA \citep{medqa}, PubMedQA \citep{pubmedqa}, BioASQ \citep{bioasq}, and MMLU-Medical \citep{hendrycks2021mmlu} collectively contain over 480K questions assessing clinical knowledge, diagnostic reasoning, and literature comprehension. The rare-disease-focused RareBench \citep{rarebench2024} adds differential-diagnosis coverage across four dimensions (phenotype extraction, screening, differential diagnosis, single-shot diagnosis). None of these benchmarks evaluates \emph{temporal} clinical reasoning. For example, questions such as ``at what age does phenotype X emerge in disease Y?'' or ``which disease has an earlier onset?'' are generally not represented in these benchmarks. GA4GH Phenopackets \citep{phenopackets} provides 6{,}668 patient-level cases with phenotype onset ages, which can serve as ground truth but is not structured as a QA benchmark. ChronoTQA addresses this gap with 3{,}329 questions across eight reported task types (plus a 12-question supplementary diagnostic set), including temporal window identification, temporal differential diagnosis, cross-disease onset comparison, and phenotype ordering. These tasks require temporal knowledge that is not systematically captured by existing KGs.

\section{ChronoMedKG Construction}
\label{sec:construction}

ChronoMedKG is built end-to-end from a disease identifier through four cooperating agents, with no per-disease code, manual curation, or prompt engineering. Two design choices distinguish the pipeline: temporal information is collected at extraction time (not as a post-hoc enrichment), and every triple carries its evidence (PMID, verbatim quote, six-signal credibility score) into the released file.

\subsection{Four-Agent Autonomous Pipeline}

ChronoMedKG is built using a multi-stage pipeline (Figure~\ref{fig:pipeline}) that processes each disease independently. Given a disease identifier (e.g., MONDO, OMIM, ORPHA), four agents execute sequentially. (1)~\textbf{Disease Profiler} queries biomedical ontologies (MONDO, OMIM, HPO) to generate disease-specific metadata, including synonyms, differential diagnoses, associated genes, PubMed counts, and an estimate of literature coverage (Standard $\geq$100 articles, Light 20--99, Minimal $<$20). (2)~\textbf{Evidence Harvester} retrieves abstracts and full-text articles from PubMed and PMC using NCBI E-utilities~\citep{sayers2022ncbi}; Standard-tier diseases get up to 150 documents, light/minimal tiers get all available literature. (3)~\textbf{Knowledge Extractor} processes documents in parallel with 2--3 LLMs from different families (DeepSeek~V3, GPT-4o-mini or GPT-4.1-nano, Claude~3 Haiku as conditional tiebreaker). Candidate triples are retained only when independently extracted by at least $\geq$2 models from the same document with entity fuzzy matching $\geq$80\% and same canonical relation. From 13{,}045{,}687 raw extractions, 460{,}497 pass consensus (3.53\%; retention-rate decomposition in Appendix~\ref{app:consensus}). (4)~\textbf{Quality Controller} aligns extracted entities to the PrimeKG schema, applies temporal plausibility constraints (e.g., age within $\in [0, 120]$), detects conflicts, and computes six-signal credibility scores (journal tier, citation velocity, study type, replication, retraction, and LLM consensus). Full per-agent prompts, model parameters, and tier definitions are documented in Appendix~\ref{app:prompts}.

\subsection{Schema and Full Evidence Grounding}
\label{sec:schema}

Each ChronoMedKG edge carries three classes of metadata. \textbf{Temporal metadata} records when phenotypes manifest, in one of three forms: onset age range (present on 24.5\% of validated triples), progression stage (16.4\%), or clinical milestone (45.5\%); a single edge may carry more than one form. \textbf{Evidence provenance} records PMIDs, the verbatim source quote ($\leq$300 chars), publication year, study type, and a six-signal credibility score. \textbf{Consensus confidence} records the fraction of extracting models that agreed (1.00 = all, 0.67 = 2/3).

\paragraph{Full evidence grounding at scale.} In contrast to prior resources, such as  PrimeKG (zero edge-level PMIDs) or iKraph (partial document traces, no quotes), ChronoMedKG provides \emph{both} the raw pool and the validated output: Among the raw extractions, 90.6\% of 13M raw triples are associated with a PMID and 99.9\% include a supporting verbatim quote; for the validated subset, all triples are linked to a PMID and 99.9\% include a corresponding quote. Multi-source triples (1.6\%, cross-paper validated) have up to 81 supporting PMIDs, enabling direct verification of the underlying evidence.

A complete example record (DMD $\rightarrow$ dilated cardiomyopathy, onset 10--18y, non-ambulatory stage) and a visualisation of DMD's full staged temporal profile are in Appendix~\ref{app:schema-example}.

\subsection{Resource Statistics}

\begin{table}[h]
\centering
\small
\caption{ChronoMedKG resource (code-verified). Both raw and validated tiers are evidence-grounded with PMID + verbatim quote.}
\label{tab:resource}
\begin{tabular}{lrl}
\toprule
\textbf{Metric} & \textbf{Value} & \textbf{Notes} \\
\midrule
\multicolumn{3}{l}{\textit{Scale}} \\
Diseases processed & 13{,}431 & 78.6\% of 17{,}080 PrimeKG diseases \\
Diseases with validated output & 10{,}852 & Validated consensus triples retained \\
Documents processed & $\sim$2.5M & 43K full-text PMC + 2.4M abstracts \\
\midrule
\multicolumn{3}{l}{\textit{Raw extraction pool (pre-consensus) - evidence-grounded at scale}} \\
Raw triples extracted & 13{,}045{,}687 & From multi-LLM extraction \\
Raw triples with PMID & 11{,}819{,}561 & \textbf{90.6\%} of all raw triples \\
Raw triples with evidence quote & 13{,}035{,}195 & \textbf{99.9\%} of all raw triples \\
Fully grounded raw triples (PMID + quote) & 11{,}809{,}565 & \textbf{90.5\%} of all raw triples \\
\midrule
\multicolumn{3}{l}{\textit{Consensus filtering (denominator: 10{,}852 diseases with validated output)}} \\
Validated consensus triples & 460{,}497 & 3.53\% of raw triples \\
3-model consensus (diseases) & 6{,}551 & 60.4\% of 10{,}852 \\
2-model consensus (diseases) & 4{,}246 & 39.1\% of 10{,}852 \\
Other configurations (4 models or 1+unlabelled) & 55 & 0.5\% of 10{,}852 \\
Full agreement (triples) & 87.1\% & All models that processed doc agree \\
Partial agreement (triples) & 12.9\% & 2/3 models agree \\
\midrule
\multicolumn{3}{l}{\textit{Validated triples - temporal metadata and full grounding}} \\
Triples with onset ages & 112{,}932 & 24.5\% \\
Triples with progression stages & 75{,}723 & 16.4\% \\
Triples with milestones & 209{,}485 & 45.5\% \\
Diseases with temporal data & 8{,}935 & 82.3\% of validated output \\
Triples with PMID citation & 460{,}497 & \textbf{100\%} (full provenance) \\
Triples with evidence text & 459{,}988 & \textbf{99.9\%} (verbatim quote $\leq$300 chars) \\
Multi-source triples ($\geq$2 PMIDs) & 7{,}563 & 1.6\% (cross-paper validation) \\
\bottomrule
\end{tabular}
\end{table}

\paragraph{Evidence age.} Among 455{,}519 validated triples, 98.9\% carry PMID-traceable publication dates, with a median year of 2015; of these, 27.4\% are supported by evidence from the past five years, while 24.1\% rely on studies $>$20 years ago. In contrast, static KGs do not provide edge-level publication dates. The full distribution and evidence-decay analysis are presented in Appendix~\ref{app:decay}.

\section{Validation}
\label{sec:validation}

We validate the resource in three steps that match the structure of the contribution. Step one re-verifies the novel-coverage subset (the diseases where no external gold standard exists) using a three-LLM judge panel against the underlying evidence text. Step two cross-checks against three external onset databases (Orphadata, HPOA, GeneReviews) on diseases that overlap with each. Step three reports the HEG-TKG clinician anchor on the underlying extraction methodology.

\subsection{Novel-Coverage Verification via Multi-LLM Judge Panel}
\label{sec:novelty}

For 6{,}250 diseases ChronoMedKG provides onset ranges absent from any curated gold standard (Section~\ref{sec:coverage}). With no external reference available, we audit the evidence-to-claim fidelity of a stratified sample: do the verbatim quotes attached to each triple actually support the claimed numeric onset range? This is a text-grounding check, not an independent re-validation of the underlying biology, and we report it as such.

We sampled $n{=}100$ diseases (seed$=$42), stratified by literature tier (Standard / Light / Minimal, defined in Section~\ref{sec:construction}) and onset bucket (clinical era: prenatal, neonatal, infantile, childhood, adolescent, adult), selecting one (claim, evidence) pair per disease. Three judges (DeepSeek~V3, GPT-4o-mini, Claude Haiku~4.5, temperature~0.0) independently rated each pair under a chain-of-thought prompt that required verbatim-quote$\to$numeric-range translation via a fixed clinical-era lookup and returned {\scshape supported} / {\scshape partially\_supported} / {\scshape not\_supported} / {\scshape unverifiable}. Full sampling, prompt, and error-taxonomy detail in Appendix~\ref{app:novelty-detail}.

\begin{table}[h]
\centering
\small
\caption{Three-LLM judge panel on $n{=}100$ novel-coverage diseases. \emph{Verifiable majority} excludes {\scshape unverifiable} and three-way splits.}
\label{tab:novelty}
\setlength{\tabcolsep}{5pt}
\begin{tabular}{lrrrr|r}
\toprule
\textbf{Verdict} & \textbf{DeepSeek V3} & \textbf{GPT-4o-mini} & \textbf{Claude H-4.5} & & \textbf{Majority} \\
\midrule
{\scshape supported}           & 80 & 70 & 75 & & 76 \\
{\scshape partially\_supported} & 2 & 11 & 10 & & 4 \\
{\scshape not\_supported}       & 14 & 7 & 12 & & 11 \\
{\scshape unverifiable}         & 4 & 12 & 3  & & 3 \\
\textit{Three-way split}        & --- & --- & --- & & 6 \\
\midrule
\multicolumn{5}{l}{\textbf{Inter-judge agreement}: 66/100 unanimous, 28/100 two-of-three, 6/100 three-way split.} \\
\multicolumn{5}{l}{\textbf{Verified accuracy} (({\scshape supported}$+${\scshape partially})/verifiable majority): $80/91=\mathbf{87.9\%}$.} \\
\bottomrule
\end{tabular}
\end{table}

All 11 majority-{\scshape not\_supported} cases share one failure mode: the extraction preserved a qualitative qualifier (e.g., ``elderly onset'', ``mid-trimester'') but did not populate numeric \texttt{onset\_age\_\{min,max\}}; a qualifier-to-range fallback would reclassify 10 of 11 (genuine noise rate 1/100 = 1\%). The 87.9\% figure applies to the novel-coverage subset only and is not a re-estimate of the 92.7\% Orphadata accuracy below.

\subsection{Cross-Gold-Standard Accuracy}
Where ChronoMedKG diseases overlap with external onset databases, we run a direct comparison: for each matched disease, we compare ChronoMedKG's median per-phenotype onset range against the gold standard's disease-level range, and a ChronoMedKG range is \emph{contained} if it falls within the gold standard. We test this on three sources (Orphadata, HPOA, and GeneReviews) with results in Table~\ref{tab:validation}. The novel-coverage audit above complements this by handling diseases without any external ground truth.

\begin{table}[h]
\centering
\small
\caption{Validation against external gold standards. Effective accuracy = strict + clinically defensible (\S\ref{sec:error-taxonomy}).}
\label{tab:validation}
\begin{tabular}{lrrr}
\toprule
\textbf{Gold Standard} & \textbf{Matched Diseases} & \textbf{Strict Precision} & \textbf{Effective Accuracy} \\
\midrule
Orphadata & 2{,}563 & 50.1\% & \textbf{92.7\%} \\
HPOA & 365 & 18.6\% & \textbf{92.6\%} \\
GeneReviews & 116 & 56.9\% & \textbf{86.2\%} \\
\bottomrule
\end{tabular}
\end{table}

\subsection{Strict Error Taxonomy: Granularity vs.\ Genuine Errors}
\label{sec:error-taxonomy}
The gap between strict precision and effective accuracy reflects granularity mismatches rather than errors. HPOA stores onset as broad categorical bins (infantile, juvenile, adult), while ChronoMedKG reports per-phenotype age ranges in years. Strict containment between these formats fails by construction in most cases, even when both sources agree on the clinical timeline; the 18.6\% strict / 92.6\% effective gap on HPOA reflects this mismatch, not extraction error. Strict precision across the three gold standards lines up with gold-standard resolution: HPOA (18.6\%) $<$ Orphadata (50.1\%) $<$ GeneReviews (56.9\%). We manually classified all 2{,}563 ChronoMedKG--Orphadata disease matches by the relationship between the ChronoMedKG onset range and the Orphadata reference range (Table~\ref{tab:errors}), and find that most discrepancies trace to ChronoMedKG's finer granularity, namely per-phenotype onset ranges where gold standards report a single disease-level bin, rather than to extraction errors. Section~\ref{sec:coverage} quantifies the granularity gap across the full resource; a stacked-bar visualisation of the error categories is in Appendix~\ref{app:error-fig}.

\begin{table}[h]
\centering
\small
\setlength{\tabcolsep}{4pt}
\caption{Error taxonomy (Orphadata, code-verified). Only 7.3\% are genuine errors.}
\label{tab:errors}
\begin{tabular}{llrr}
\toprule
\textbf{Category} & \textbf{Description} & \textbf{\%} & \textbf{Verdict} \\
\midrule
Contained (correct) & Range within gold standard & 50.1\% & Correct \\
Adjacent stage & Differs from gold by one clinical era & 15.6\% & Not an error \\
Granularity mismatch & Per-phenotype vs gold disease-level bin & 13.8\% & Not an error \\
Wider but overlaps & Captures early$+$late phenotypes; core overlaps & 6.7\% & Not an error \\
Single-triple noise & Outlier triple pulls range; median correct & 5.7\% & Not an error \\
\textbf{Genuinely wrong} & \textbf{No overlap, $>$10y gap} & \textbf{7.3\%} & \textbf{Real error} \\
\bottomrule
\end{tabular}
\end{table}

\subsection{Clinician Anchor (via HEG-TKG)}
\label{sec:clinician}
On three diagnostic pairs of rare neuromuscular diseases (DMD/BMD, MG/LEMS, CIDP/GBS), correct diagnosis hinges on temporal cues such as age of onset, progression rate, or symptom duration. A three-rater clinician panel evaluated our system against vanilla GPT-4.1. The senior neurologist (C1, 35 blinded cases) rated ChronoMedKG markedly more verifiable ($\Delta{=}{+}1.64/5$, $d{=}1.79$, $p{<}0.001$, BH-corrected); a second senior neurologist replicated this on the 11-case CIDP/GBS subset with the largest effect in the panel ($d{=}2.57$, $p{<}0.003$); a trainee returned a smaller but still BH-significant advantage ($d{=}0.67$, $q{=}0.006$), noting that our system cites consensus knowledge specialists already know but non-specialists need.

This evaluation covers a small subset of diseases under a simplified extraction protocol (titles and abstracts only); clinician review at ChronoMedKG scale across the 6{,}250 novel-coverage diseases is the remaining step (Limitation~(ii), Section~\ref{sec:discussion}). Full Likert breakdowns by rater and dimension in Appendix~\ref{app:clinician}.

\section{ChronoTQA Benchmark}
\label{sec:benchmark}
On the basis of ChronoMedKG, we introduce \textbf{ChronoTQA}, the first temporal biomedical QA benchmark. The benchmark contains 3{,}341 questions: 3{,}329 across eight reported task types plus a 12-question HPOA-grounded negative-temporal MCQ probe held back as a supplementary diagnostic set (because $n{=}12$ is too small for per-type leaderboard reporting). ChronoTQA tests \emph{when} phenotypes emerge in the course of a disease; it complements rare-disease differential-diagnosis benchmarks like RareBench \citep{rarebench2024} that test \emph{which} disease a phenotype set indicates. None of the existing biomedical QA benchmarks discussed in \S\ref{sec:related} (MedQA, PubMedQA, BioASQ, MMLU-Medical, RareBench) evaluates age- or stage-conditioned reasoning; a side-by-side landscape comparison is in supplementary Appendix~D.1. Table~\ref{tab:tqa} reports the ChronoTQA composition.

\paragraph{ChronoTQA grows with its sources.} Both tiers expand with their underlying data. Tier~1 questions regenerate when Orphadata, HPOA, or Phenopackets release new versions, and the generation scripts ship with the benchmark. Tier~2 items trace to versioned ChronoMedKG triples and grow with each release, with full PMID provenance preserved. The benchmark scales with the underlying clinical literature; v1.0 is the first release in a versioned series, not a fixed snapshot.

\begin{table}[h]
\centering
\small
\caption{ChronoTQA composition. Tier~1 grounds in external databases; Tier~2 traces to PMID-verified ChronoMedKG triples; static controls sanity-check basic biomedical recall. The 12-question HPOA negative-temporal MCQ probe is included in the release as a supplementary diagnostic set; $n{=}12$ is too small to support per-type leaderboard claims, so we exclude it from the per-type reporting.}
\label{tab:tqa}
\setlength{\tabcolsep}{4pt}
\begin{tabular}{llrll}
\toprule
\textbf{Tier} & \textbf{Type} & \textbf{N} & \textbf{Probe} & \textbf{Gold Source} \\
\midrule
\multirow{4}{*}{Tier 1 (external)}
 & Temporal window & 800 & Phenotype-onset age & Orphadata \\
 & Temporal differential Dx & 687 & Disease from onset profile & Orphadata \\
 & Cross-disease comparison & 600 & Earlier-onset disease & Orphadata \\
 & Phenopackets onset & 147 & Precise age from case data & Phenopackets \\
\midrule
\multirow{2}{*}{Tier 2 (KG-derived)}
 & Phenotype ordering & 395 & Does $A$ precede $B$ in $D$? & ChronoMedKG \\
 & Stage-conditional & 200 & Stage-specific features & ChronoMedKG \\
\midrule
\multirow{2}{*}{Static controls}
 & Drug--disease indication & 250 & Indication (control) & PrimeKG \\
 & Gene--disease association & 250 & Association (control) & PrimeKG \\
\midrule
\multicolumn{2}{l}{\textbf{Subtotal (8 task types)}} & \textbf{3{,}329} & & \\
\multicolumn{2}{l}{Supplementary HPOA negative-temporal MCQ probe} & 12 & Rule-out by age & HPOA \\
\multicolumn{2}{l}{\textbf{Total release}} & \textbf{3{,}341} & & \\
\bottomrule
\end{tabular}
\end{table}

\paragraph{Quality assurance.} Four rounds of multi-LLM independent evaluation (540 assessments across 180 questions, 3 evaluator LLMs per question) verified well-formedness and answer correctness. Automated QC removed 30 logically-inconsistent questions. All 1{,}447 external-source answers verified against source data with 0 mismatches.

\paragraph{Frontier-LLM baseline: where the difficulty concentrates.}
\label{sec:llm-baseline}
We drew a stratified 120-question subsample from the 3{,}329 reported items and posed it to four frontier LLMs via web chat (no API, no system prompts, no retrieval): GPT-4o-mini, Gemini, Claude, and DeepSeek V3. Per-model answer logs ship with the benchmark release (Appendix~\ref{app:llm_gap}). Table~\ref{tab:llm_gap} shows the per-task breakdown.

The aggregate gap is $+$30.1\,pp (static minus temporal, mean across models); the breakdown shows where it concentrates. Binary cross-disease comparison and the static controls saturate near ceiling. Bounded MCQ subtasks (temporal window, temporal differential diagnosis) sit in the 36--83\% range and split unevenly across models. The free-text Phenopackets onset subtask, which asks for a precise age range without multiple-choice scaffolding, collapses to a 4.5\% mean across models (per-model rates 0--12\%). Frontier LLMs cannot produce calibrated age windows from parametric memory. Section~\ref{sec:rag} reports retrieval against ChronoMedKG that recovers 47--65\% of the questions each model fails without retrieval on the Phenopackets-grounded onset subset; PrimeKG-RAG recovers 40--51\%, HPOA-RAG 17--29\%. Full per-model breakdowns are in Appendix~\ref{app:llm_gap}.

\begin{table}[h]
\centering
\small
\caption{Per-task ChronoTQA accuracy on a 120-question stratified subsample (4 frontier LLMs, web chat, no retrieval). Free-text precise-age collapses near zero; static controls and binary comparisons saturate. Tier 2 types reported in \S\ref{sec:rag} and Appendix~\ref{app:llm_gap}.}
\label{tab:llm_gap}
\setlength{\tabcolsep}{5pt}
\begin{tabular}{lcccc|c}
\toprule
\textbf{Task} & \textbf{GPT-4o-mini} & \textbf{Gemini} & \textbf{Claude} & \textbf{DeepSeek} & \textbf{Mean} \\
\midrule
Cross-disease comparison       & 100\% & 100\% & 100\% & \phantom{0}91\% & 98\% \\
Temporal window                & \phantom{0}58\%  & \phantom{0}83\%  & \phantom{0}75\%  & \phantom{0}67\% & 71\% \\
Temporal differential Dx       & \phantom{0}82\%  & \phantom{0}75\%  & \phantom{0}55\%  & \phantom{0}36\% & 62\% \\
Phenopackets onset (free-text) & \phantom{00}0\%  & \phantom{0}12\%  & \phantom{00}0\%  & \phantom{00}6\% & \textbf{\phantom{0}4.5\%} \\
\midrule
Static drug (control)          & \phantom{0}80\%  & 100\% & \phantom{0}80\%  & \phantom{0}80\% & 85\% \\
Static gene (control)          & \phantom{0}75\%  & 100\% & \phantom{0}75\%  & \phantom{0}75\% & 81\% \\
\midrule
\textit{Per-model temporal mean}  & \textit{52.9\%} & \textit{64.0\%} & \textit{51.0\%} & \textit{45.1\%} & - \\
\textit{Per-model static mean}    & \textit{77.8\%} & \textit{100.0\%} & \textit{77.8\%} & \textit{77.8\%} & - \\
\textit{Gap (static $-$ temporal, pp)} & \textit{$+$24.8} & \textit{$+$36.0} & \textit{$+$26.8} & \textit{$+$32.7} & \textit{$+$30.1} \\
\bottomrule
\end{tabular}
\end{table}

\section{Novel Analyses Enabled by Temporal Grounding}
\label{sec:novel}

Two analyses use the temporal annotations directly: a coverage-gap study and a retrieval experiment on long-tail questions. Link prediction, evidence decay, trajectory clustering, and bin-granularity sensitivity are in Supplementary Appendices~H.7, H.8, I.1, and~I.4. 

\subsection{Coverage Gap: What Temporal Data Exists Where}
\label{sec:coverage}

We evaluate coverage by testing whether onset age information is available in existing structured biomedical resources for each of the 17{,}080 diseases in PrimeKG. Table~\ref{tab:coverage} shows the result. PrimeKG does not contain temporal metadata and therefore provides no onset information. HPOA covers 8.4\% of diseases, Orphadata 33.9\%, and Phenopackets 3.0\%. In comparison, ChronoMedKG provides onset information for 52.3\% of diseases. \textbf{6{,}250 diseases} (36.6\% of all PrimeKG diseases) have onset data that is not present in any reference resource.

\begin{table}[h]
\centering
\small
\caption{Onset-age coverage vs.\ PrimeKG's 17{,}080 diseases. Granularity = per-phenotype or disease-level.}
\label{tab:coverage}
\begin{tabular}{lrrl}
\toprule
\textbf{Resource} & \textbf{Diseases with onset} & \textbf{Coverage} & \textbf{Granularity} \\
\midrule
PrimeKG & 0 & 0.0\% & None \\
HPOA & 1{,}429 & 8.4\% & Coarse bins (disease-level) \\
Phenopackets & 518 & 3.0\% & Per-patient (case data) \\
Orphadata & 5{,}796 & 33.9\% & Coarse bins (disease-level) \\
\textbf{ChronoMedKG} & \textbf{8{,}935} & \textbf{52.3\%} & \textbf{Per-phenotype, numeric} \\
\midrule
\textbf{ChronoMedKG novel coverage} & \textbf{6{,}250} & \textbf{36.6\%} & Not in any gold standard \\
\bottomrule
\end{tabular}
\end{table}

\begin{figure}[h]
    \centering
    \includegraphics[width=\linewidth]{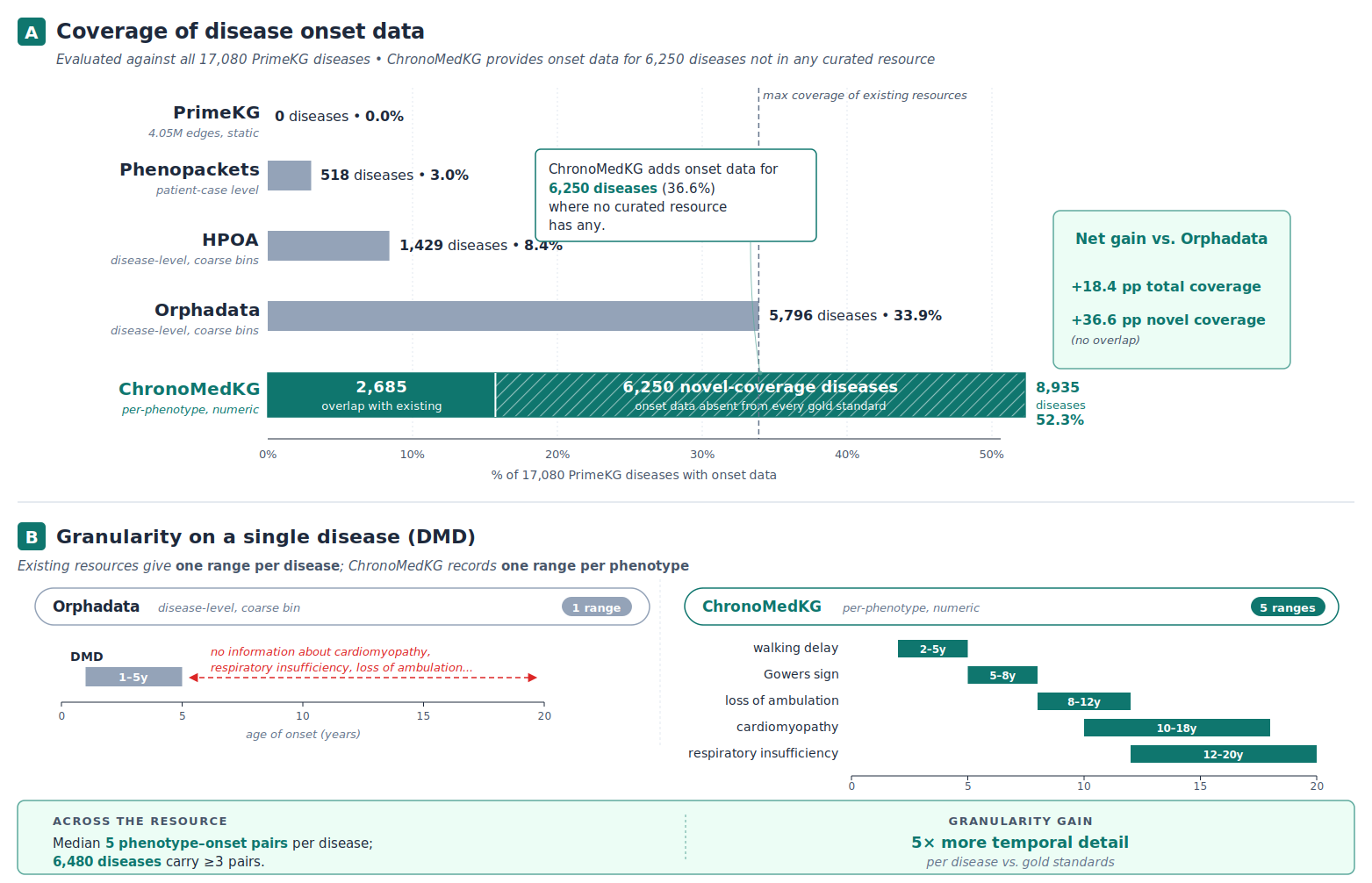}
    \caption{\textbf{Coverage and granularity of disease onset data.} \textbf{(A)}~ChronoMedKG covers 8{,}935 of 17{,}080 PrimeKG diseases (52.3\%); \textbf{6{,}250} have onset data absent from every curated gold standard ($+$18.4\,pp over Orphadata's 33.9\%). \textbf{(B)}~DMD granularity contrast: Orphadata gives a single 1--5y range; ChronoMedKG records five per-phenotype onsets (walking delay 2--5y, Gowers sign 5--8y, loss of ambulation 8--12y, cardiomyopathy 10--18y, respiratory insufficiency 12--20y). Across the resource: median 5 phenotype-onset pairs per disease ($n{=}6{,}480$ with $\geq$3), $\sim$5$\times$ more detail than any gold standard.}
    \label{fig:coverage}
\end{figure}

Beyond coverage, ChronoMedKG provides substantially finer granularity. Orphadata reports a single range per disease (e.g., ``DMD: 1--5 years''); ChronoMedKG provides a median of 5 distinct phenotype-onset pairs per disease, with 6{,}480 diseases having $\geq$3 such pairs. In DMD, for example, ChronoMedKG separately records walking delay (2--5y), Gowers sign (5--8y), loss of ambulation (8--12y), and cardiomyopathy (10--18y). This information is absent from every existing structured resource. Figure~\ref{fig:coverage} visualizes this gap. Among the 6{,}250 novel-coverage diseases, 1{,}657 are Orphanet-coded rare diseases gaining first-time temporal grounding; 4{,}593 are PrimeKG diseases unmatched by any curated onset resource.

\subsection{Retrieval-Augmented Generation: Rescue on Long-Tail Questions}
\label{sec:rag}

We evaluated whether retrieval from ChronoMedKG improves LLM performance on 147 externally-grounded GA4GH Phenopackets questions drawn from the ChronoTQA benchmark (Tier~1). These 147 questions cover the 518 diseases that Phenopackets describes, where ground truth exists. They are not the 6{,}250 novel-coverage diseases of \S\ref{sec:coverage}; in that pool, no external resource has data, so there is no way to score one retrieval source against another. The rescue experiment therefore runs on shared territory, where every retrieval source has a fair shot. Three LLMs (Claude~3 Haiku, DeepSeek~V3, GPT-4o-mini) answered each question under four conditions: no retrieval (NR), PrimeKG-RAG, HPOA-RAG, and ChronoMedKG-RAG, scored with a calibrated rubric. The Phenopackets corpus has been public since 2019, so LLMs reach 76--86\% accuracy from parametric memory alone (NR). This is the ceiling set by what these models already know, and all four retrieval sources operate against it.

A better test under this ceiling is \emph{long-tail rescue}: on the subset where each model fails without retrieval, how often does each retrieval source recover the correct answer? Table~\ref{tab:rag-rescue} reports the result, with 95\% bootstrap confidence intervals from 10{,}000 resamples (seed=42). ChronoMedKG rescues \textbf{47--65\%} of NR-failed questions across the three models. The improvement over HPOA-RAG (17--29\%) is robust under bootstrap resampling for two of three models (Claude, DeepSeek), where the CIs do not overlap. The improvement over PrimeKG-RAG (40--51\%) holds at the point estimate; CIs overlap given the n=20--35 NR-fail subset per model. ChronoMedKG is the best rescue source at the point estimate for all three models tested.

Pooled accuracy is bounded by the parametric ceiling: ChronoMedKG-RAG reaches 82.1\% vs.\ 79.8\% NR ($+$2.3\,pp), with Claude~3 Haiku the only per-model significance (McNemar $p{=}0.008$, $+$12.9\,pp). Full leaderboard and per-model significance are in Appendix~\ref{app:rag}. Retrieval is one strategy for KG-grounded LLM use; runtime triple verification (KGARevion \citep{kgarevion2025}) is an orthogonal one, enabled by grounded temporal KGs like ChronoMedKG.

\begin{table}[h]
\centering
\small
\caption{Long-tail rescue on 147 Phenopackets-grounded onset questions (ChronoTQA Tier~1). Fraction of NR-failed questions each retrieval source recovers per model. Brackets are 95\% bootstrap CIs over 10{,}000 resamples (seed=42). ChronoMedKG is the best rescue source at the point estimate for all three models; the gain over HPOA-RAG is statistically robust for Claude and DeepSeek.}
\label{tab:rag-rescue}
\begin{tabular}{lrrrr}
\toprule
\textbf{Model} & \textbf{NR-fail $n$} & \textbf{PrimeKG rescues} & \textbf{HPOA rescues} & \textbf{ChronoMedKG rescues} \\
\midrule
Claude~3 Haiku & 35 & 51.4\% [34, 69] & 17.1\% [6, 31] & \textbf{60.0\% [43, 77]} \\
DeepSeek~V3    & 20 & 40.0\% [20, 60] & 20.0\% [5, 40] & \textbf{65.0\% [45, 85]} \\
GPT-4o-mini    & 34 & 41.2\% [26, 59] & 29.4\% [15, 44] & \textbf{47.1\% [29, 65]} \\
\bottomrule
\end{tabular}
\end{table}

\paragraph{Clinical-case illustration.}
\label{sec:cases}
On 31 open-access PMC diagnostic-odyssey case reports (delays 1--50\,y, median 14, 11 disease categories), we matched observed phenotype sequences against ChronoMedKG's staged profiles. In PMC5688900, 26 years of schizophrenia misdiagnosis preceded a Wilson-disease diagnosis; ChronoMedKG records neuropsychiatric onset at 20--40\,y alongside hepatic involvement at 15--45\,y, the timing pattern that separates Wilson from primary psychiatric illness. Full 31-case table, timeline, and per-case temporal-discriminator analysis are in Appendix~\ref{app:pmc-cases}.

\section{Discussion}
\label{sec:discussion}

ChronoMedKG complements three concurrent agentic biomedical KG systems along an axis the others do not touch. KARMA~\citep{karma2025} resolves conflicts between multi-agent extractors. MedKGent~\citep{medkgent2025} tracks how knowledge enters the literature over publication time. AutoBioKG~\citep{autobiokg2026} encodes contextual state. None of these encodes clinical time, the patient-age and disease-stage axis on which clinical decisions actually get made. ChronoMedKG fills that gap at PrimeKG scale, with edge-level onset ages, progression stages, and disease-course milestones grounded to PMIDs.

Two findings carry the resource, and the populations behind them are different. First, 6{,}250 diseases gain temporal grounding that no other resource provides (\S\ref{sec:coverage}); 1{,}657 of those are Orphanet-coded rare diseases getting structured onset for the first time. Second, on long-tail onset questions where Phenopackets supplies the ground truth, ChronoMedKG retrieval rescues 47--65\% of the queries each frontier LLM answers incorrectly without retrieval, against 40--51\% for static PrimeKG and 17--29\% for HPOA (\S\ref{sec:rag}). These are not the same diseases. The first finding is that ChronoMedKG has data that no other source does; the second is that ChronoMedKG is more precise in the territory all four sources share.

\paragraph{Limitations.} Six gaps that ChronoMedKG has, in descending order of how much they constrain near-term use.
(i)~\textit{Entity canonicalisation}: LLM-extracted phenotype names give ChronoMedKG a 3$\times$ larger entity space than HPOA, suppressing link-prediction MRR; SapBERT~\citep{liu2021sapbert}.
(ii)~\textit{Clinician validation at scale}: the six-disease HEG-TKG panel (\S\ref{sec:clinician}) anchors the methodology, not the full 13{,}431-disease resource; the 6{,}250 novel-coverage diseases need expert review at scale.
(iii)~\textit{Extraction error}: $\geq$2-model consensus still leaves 7.3\% genuinely wrong (Table~\ref{tab:errors}), which limits single-triple reliance and favours aggregate or median-based use.
(iv)~\textit{Deep tier never triggered}: all triples are PubMed-extracted; GeneReviews, OMIM, and other databases are not yet integrated.
(v)~\textit{Credibility score}: two of six signals (\texttt{citation\_count}, \texttt{is\_retracted}) are unpopulated.
(vi)~\textit{Research scope}: literature-derived; direct clinical use requires clinician oversight (Ethics, Appendix~\ref{app:ethics}).


\bibliographystyle{plainnat}
\bibliography{references}

@article{chandak2023primekg,
  title={Building a knowledge graph to enable precision medicine},
  author={Chandak, Payal and Huang, Kexin and Zitnik, Marinka},
  journal={Scientific Data},
  volume={10},
  number={1},
  pages={67},
  year={2023},
  publisher={Nature Publishing Group},
  doi={10.1038/s41597-023-01960-3}
}

@article{himmelstein2017hetionet,
  title={Systematic integration of biomedical knowledge prioritizes drugs for repurposing},
  author={Himmelstein, Daniel Scott and Lizee, Antoine and Hessler, Christine and Brueggeman, Leo and Chen, Sabrina L and Hadley, Dexter and Green, Ari and Khankhanian, Pouya and Baranzini, Sergio E},
  journal={eLife},
  volume={6},
  pages={e26726},
  year={2017},
  doi={10.7554/eLife.26726}
}

@article{su2025ikraph,
  title={A comprehensive large-scale biomedical knowledge graph for {AI}-powered data-driven biomedical research},
  author={Zhang, Yuan and Sui, Xin and Pan, Feng and Yu, Kaixian and Li, Keqiao and Tian, Shubo and Erdengasileng, Arslan and Han, Qing and Wang, Wanjing and Wang, Jianan and Wang, Jian and Sun, Donghu and Chung, Henry and Zhou, Jun and Zhou, Eric and Lee, Ben and Zhang, Peili and Qiu, Xing and Zhao, Tingting and Zhang, Jinfeng},
  journal={Nature Machine Intelligence},
  volume={7},
  pages={602--614},
  year={2025},
  doi={10.1038/s42256-025-01014-w}
}

@inproceedings{karma2025,
  title={{KARMA}: Leveraging Multi-Agent {LLMs} for Automated Knowledge Graph Enrichment},
  author={Lu, Yuxing and Wu, Wei and Zhao, Xukai and Peng, Rui and Wang, Jinzhuo},
  year={2025},
  booktitle={Advances in Neural Information Processing Systems (NeurIPS)},
  volume={38},
  note={NeurIPS 2025 Spotlight Poster. arXiv:2502.06472}
}

@article{hpoa,
  title={The {Human Phenotype Ontology} in 2024: phenotypes around the world},
  author={K\"{o}hler, Sebastian and Gargano, Michael and Matentzoglu, Nicolas and others},
  journal={Nucleic Acids Research},
  volume={52},
  number={D1},
  pages={D1333--D1346},
  year={2024},
  doi={10.1093/nar/gkad1005}
}

@article{phenopackets,
  title={The {GA4GH} {Phenopacket} schema defines a computable representation of clinical data},
  author={Jacobsen, Julius O B and Baudis, Michael and Baynam, Gareth S and others},
  journal={Nature Biotechnology},
  volume={40},
  number={6},
  pages={817--820},
  year={2022},
  doi={10.1038/s41587-022-01357-4}
}

@article{medqa,
  title={What Disease does this Patient Have? {A} Large-scale Open Domain Question Answering Dataset from Medical Exams},
  author={Jin, Di and Pan, Eileen and Oufattole, Nassim and Weng, Wei-Hung and Fang, Hanyi and Szolovits, Peter},
  journal={Applied Sciences},
  volume={11},
  number={14},
  pages={6421},
  year={2021},
  doi={10.3390/app11146421}
}

@inproceedings{pubmedqa,
  title={{PubMedQA}: A Dataset for Biomedical Research Question Answering},
  author={Jin, Qiao and Dhingra, Bhuwan and Liu, Zhengping and Cohen, William W and Lu, Xinghua},
  booktitle={Proceedings of the 2019 Conference on Empirical Methods in Natural Language Processing (EMNLP)},
  pages={2567--2577},
  year={2019},
  doi={10.18653/v1/D19-1259}
}

@article{bioasq,
  title={An overview of the {BIOASQ} large-scale biomedical semantic indexing and question answering competition},
  author={Tsatsaronis, George and Balikas, Georgios and Malakasiotis, Prodromos and others},
  journal={BMC Bioinformatics},
  volume={16},
  pages={138},
  year={2015},
  doi={10.1186/s12859-015-0564-6}
}

@inproceedings{transe,
  title={Translating Embeddings for Modeling Multi-relational Data},
  author={Bordes, Antoine and Usunier, Nicolas and Garc\'{i}a-Dur\'{a}n, Alberto and Weston, Jason and Yakhnenko, Oksana},
  booktitle={Advances in Neural Information Processing Systems},
  volume={26},
  year={2013}
}

@inproceedings{trivedi2017knowevolve,
  title={{Know-Evolve}: Deep Temporal Reasoning for Dynamic Knowledge Graphs},
  author={Trivedi, Rakshit and Dai, Hanjun and Wang, Yichen and Song, Le},
  booktitle={Proceedings of the 34th International Conference on Machine Learning (ICML)},
  pages={3462--3471},
  year={2017}
}

@inproceedings{dasgupta2018hyte,
  title={{HyTE}: Hyperplane-based Temporally aware Knowledge Graph Embedding},
  author={Dasgupta, Shib Sankar and Ray, Swayambhu Nath and Talukdar, Partha},
  booktitle={Proceedings of the 2018 Conference on Empirical Methods in Natural Language Processing (EMNLP)},
  pages={2001--2011},
  year={2018},
  doi={10.18653/v1/D18-1225}
}

@inproceedings{t_phenotype2023,
  title={{T-Phenotype}: Discovering Phenotypes of Predictive Temporal Patterns in Disease Progression},
  author={Qin, Yuchao and van der Schaar, Mihaela and Lee, Changhee},
  booktitle={Proceedings of the 26th International Conference on Artificial Intelligence and Statistics (AISTATS)},
  pages={3466--3492},
  year={2023},
  series={Proceedings of Machine Learning Research},
  volume={206}
}

@article{gebru2021datasheets,
  title={Datasheets for datasets},
  author={Gebru, Timnit and Morgenstern, Jamie and Vecchione, Briana and Vaughan, Jennifer Wortman and Wallach, Hanna and Daum{\'e} III, Hal and Crawford, Kate},
  journal={Communications of the ACM},
  volume={64},
  number={12},
  pages={86--92},
  year={2021},
  doi={10.1145/3458723}
}

@inproceedings{liu2021sapbert,
  title={Self-Alignment Pretraining for Biomedical Entity Representations},
  author={Liu, Fangyu and Shareghi, Ehsan and Meng, Zaiqiao and Basaldella, Marco and Collier, Nigel},
  booktitle={Proceedings of the 2021 Conference of the North American Chapter of the Association for Computational Linguistics: Human Language Technologies (NAACL-HLT)},
  pages={4228--4238},
  year={2021},
  doi={10.18653/v1/2021.naacl-main.334}
}

@article{sayers2022ncbi,
  title={Database resources of the national center for biotechnology information},
  author={Sayers, Eric W and Bolton, Evan E and Brister, J Rodney and Canese, Kathi and Chan, Jessica and Comeau, Donald C and Connor, Ryan and Funk, Kathryn and Kelly, Chris and Kim, Sunghwan and others},
  journal={Nucleic Acids Research},
  volume={50},
  number={D1},
  pages={D20--D26},
  year={2022},
  doi={10.1093/nar/gkab1112}
}

@inproceedings{sun2019rotate,
  title={{RotatE}: Knowledge Graph Embedding by Relational Rotation in Complex Space},
  author={Sun, Zhiqing and Deng, Zhi-Hong and Nie, Jian-Yun and Tang, Jian},
  booktitle={International Conference on Learning Representations (ICLR)},
  year={2019},
  note={arXiv:1902.10197}
}

@inproceedings{hendrycks2021mmlu,
  title={Measuring Massive Multitask Language Understanding},
  author={Hendrycks, Dan and Burns, Collin and Basart, Steven and Zou, Andy and Mazeika, Mantas and Song, Dawn and Steinhardt, Jacob},
  booktitle={International Conference on Learning Representations (ICLR)},
  year={2021},
  note={arXiv:2009.03300}
}

@article{ali2021pykeen,
  title={{PyKEEN 1.0}: A {Python} Library for Training and Evaluating Knowledge Graph Embeddings},
  author={Ali, Mehdi and Berrendorf, Max and Hoyt, Charles Tapley and Vermue, Laurent and Sharifzadeh, Sahand and Tresp, Volker and Lehmann, Jens},
  journal={Journal of Machine Learning Research},
  volume={22},
  number={82},
  pages={1--6},
  year={2021},
  note={arXiv:2007.14175}
}

@article{ahmed2026hegtkg,
  title={The Provenance Gap in Clinical {AI}: Evidence-Traceable Temporal Knowledge Graphs for Rare Disease Reasoning},
  author={Ahmed, Md Shamim and Dusanic, Maja and Kirschner, Moritz Nikolai and Nyoungui, Elisabeth and Zsch{\"u}ntzsch, Jana and Galke Poech, Lukas and R{\"o}ttger, Richard},
  journal={arXiv preprint arXiv:2604.17114},
  year={2026},
  month={Apr},
  eprint={2604.17114},
  archivePrefix={arXiv},
  primaryClass={cs.CL},
  doi={10.48550/arXiv.2604.17114},
  url={https://arxiv.org/abs/2604.17114}
}

@article{medkgent2025,
  title         = {{MedKGent}: A Large Language Model Agent Framework for Constructing Temporally Evolving Medical Knowledge Graph},
  author        = {Zhang, Duzhen and Wang, Zixiao and Li, Zhong-Zhi and Yu, Yahan and Jia, Shuncheng and Dong, Jiahua and Xu, Haotian and Wu, Xing and Zhang, Yingying and Zhang, Tielin and Yang, Jie and Chen, Xiuying and Song, Le},
  journal       = {arXiv preprint arXiv:2508.12393},
  year          = {2025},
  doi           = {10.48550/arXiv.2508.12393},
  eprint        = {2508.12393},
  archivePrefix = {arXiv},
  primaryClass  = {cs.CL},
  url           = {https://arxiv.org/abs/2508.12393}
}

@article{autobiokg2026,
  title   = {Automating Biomedical Knowledge Graph Construction For Context-Aware Scientific Inference},
  author  = {Zheng, Yikai and Liu, Wanquan and Zeng, Bi and Feng, Yichun and Du, Xiawei and Zhou, Lu and Li, Yixue},
  journal = {bioRxiv},
  year    = {2026},
  doi     = {10.64898/2026.01.14.699420},
  url     = {https://www.biorxiv.org/content/10.64898/2026.01.14.699420v1},
  note    = {Preprint, posted 14 January 2026}
}

@inproceedings{rarebench2024,
  title     = {{RareBench}: Can {LLMs} Serve as Rare Diseases Specialists?},
  author    = {Chen, Xuanzhong and Mao, Xiaohao and Guo, Qihan and Wang, Lun and Zhang, Shuyang and Chen, Ting},
  booktitle = {Proceedings of the 30th ACM SIGKDD Conference on Knowledge Discovery and Data Mining (KDD '24)},
  year      = {2024},
  publisher = {ACM},
  doi       = {10.1145/3637528.3671576},
  url       = {https://doi.org/10.1145/3637528.3671576}
}

@inproceedings{kgarevion2025,
  title     = {{KGARevion}: An {AI} Agent for Knowledge-Intensive Biomedical {QA}},
  author    = {Su, Xiaorui and Wang, Yibo and Gao, Shanghua and Liu, Xiaolong and Giunchiglia, Valentina and Clevert, Djork-Arn{\'e} and Zitnik, Marinka},
  booktitle = {The Thirteenth International Conference on Learning Representations (ICLR)},
  year      = {2025},
  url       = {https://openreview.net/forum?id=tnB94WQGrn}
}

\section*{Data, Code, and AI Usage}
ChronoMedKG (CC~BY~4.0) and ChronoTQA are archived on Zenodo at \url{https://doi.org/10.5281/zenodo.19697542}; pipeline and experiment code (MIT) are on GitLab at \url{https://gitlab.sdu.dk/screen4care/chronomedkg}, with a Hugging Face mirror planned for camera-ready release. Construction used DeepSeek~V3 and OpenAI models (GPT-4.1-nano, then GPT-4o-mini), with Claude~3~Haiku as a conditional tiebreaker. The novelty-judge panel used DeepSeek~V3, GPT-4o-mini, and Claude Haiku 4.5; model IDs, settings, and prompts are in Appendix~\ref{app:prompts}. Claude (Anthropic) was used for limited editorial assistance, including language refinement and structural revision under author supervision; Grammarly and Overleaf handled copyediting. Figure~\ref{fig:pipeline}, the four-agent pipeline schematic, was created with assistance from an AI diagram-generation tool and then reviewed and refined by the authors for technical accuracy. All quantitative figures were generated directly from benchmark data using the released Python scripts. No data-bearing element in any figure was produced or altered by AI. All scientific content, experimental design, data collection, analysis, and interpretation were performed and verified by the authors, who take responsibility for the manuscript content.

\appendix

\section{NeurIPS Paper Checklist}
\label{app:checklist}

\begin{enumerate}

\item {\bf Claims}
    \item[] Question: Do the main claims made in the abstract and introduction accurately reflect the paper's contributions and scope?
    \item[] Answer: \textcolor{blue}{[Yes]}
    \item[] Justification: The abstract makes four headline claims, each supported by a numbered section in the main text: (1) 460{,}497 validated triples derived from 13{,}431 PrimeKG diseases; (2) 92.7\% effective accuracy against Orphadata with a strict 7.3\% genuine-error rate (main paper Section~4.3); (3) 87.9\% verified accuracy on the three-LLM novel-coverage audit (main paper Section~4.1); (4) selective retrieval against ChronoMedKG rescues 47--65\% of LLM long-tail failures across three frontier models, alongside a TransE link-prediction ablation that confirms the temporal annotations carry signal under standard KG evaluation (Appendix~\ref{app:linkpred-full}). All claims are cross-referenced in main paper Sections~3--6.

\item {\bf Limitations}
    \item[] Question: Does the paper discuss the limitations of the work performed by the authors?
    \item[] Answer: \textcolor{blue}{[Yes]}
    \item[] Justification: Main paper Section~7 (Discussion) lists six explicit limitations, in descending severity: (i)~entity canonicalisation (3$\times$ larger phenotype entity space than HPOA, suppresses link-prediction MRR; SapBERT slated for v1.1), (ii)~clinician validation at scale (six-disease HEG-TKG anchor only; 13{,}431-disease scale unvalidated by clinicians), (iii)~7.3\% extraction error after $\geq$2-model consensus (Table~5), (iv)~Deep tier never triggered (GeneReviews/OMIM not yet integrated), (v)~credibility score uncalibrated in v1.0 (two of six signals unpopulated; recalibration is v1.1), (vi)~research scope (literature-derived; direct clinical use requires clinician oversight). The LLM judge calibration in Appendix~\ref{app:judge} further quantifies the limitation of automated quality assessment.

\item {\bf Theory assumptions and proofs}
    \item[] Question: For each theoretical result, does the paper provide the full set of assumptions and a complete (and correct) proof?
    \item[] Answer: \textcolor{blue}{[N/A]}
    \item[] Justification: This is a dataset/benchmark paper; no theoretical results are claimed.

\item {\bf Experimental result reproducibility}
    \item[] Question: Does the paper fully disclose all the information needed to reproduce the main experimental results of the paper?
    \item[] Answer: \textcolor{blue}{[Yes]}
    \item[] Justification: Appendix~\ref{app:methods} describes the 4-agent pipeline, consensus algorithm, credibility scoring formula, and experimental setup with hyperparameters (link prediction: TransE, 100 epochs, 100-dim, lr=0.01, 3 seeds). All experiment scripts are in the released code repository.

\item {\bf Open access to data and code}
    \item[] Question: Does the paper provide open access to the data and code, with sufficient instructions to faithfully reproduce the main experimental results?
    \item[] Answer: \textcolor{blue}{[Yes]}
    \item[] Justification: Section~\ref{app:availability} provides URLs to the dataset (Zenodo DOI) and code repository (GitLab). Instructions include exact Python environment (Python 3.12), dependencies (\texttt{requirements.txt}), and random seeds for all experiments.

\item {\bf Experimental setting/details}
    \item[] Question: Does the paper specify all the training and test details necessary to understand the results?
    \item[] Answer: \textcolor{blue}{[Yes]}
    \item[] Justification: Section~\ref{app:linkpred-details} details train/val/test splits (80/10/10), hyperparameters (TransE embedding dim 100, batch size 1024, Adam lr=0.01, 100 epochs), optimizer, and random seeds (42, 7, 123).

\item {\bf Experiment statistical significance}
    \item[] Question: Does the paper report error bars suitably and correctly defined or other appropriate information about the statistical significance of the experiments?
    \item[] Answer: \textcolor{blue}{[Yes]}
    \item[] Justification: Link prediction results (Table~\ref{tab:linkpred}) report mean $\pm$ std across 3 seeds with paired $t$-test $p$-values. Validation accuracy uses exact counts (n=2{,}563 for Orphadata, n=365 for HPOA, n=116 for GeneReviews).

\item {\bf Experiments compute resources}
    \item[] Question: For each experiment, does the paper provide sufficient information on the computer resources needed to reproduce the experiments?
    \item[] Answer: \textcolor{blue}{[Yes]}
    \item[] Justification: Pipeline cost: \$2{,}400 total, \$0.18/disease (LLM API calls). Link prediction trained on Apple M4 MPS (no GPU required); each 100-epoch training takes $\sim$15 seconds. Full 12-run seed experiment completes in $\sim$3 minutes.

\item {\bf Code of ethics}
    \item[] Question: Does the research conducted in the paper conform, in every respect, with the NeurIPS Code of Ethics?
    \item[] Answer: \textcolor{blue}{[Yes]}
    \item[] Justification: The resource is built entirely from open-access biomedical literature (PubMed, PMC Open Access Subset). No personally-identifiable information is included. Case study PMC references are to published open-access clinical case reports with patient anonymity preserved in the source publications. The resource explicitly warns against use for direct patient care without clinical oversight.

\item {\bf Broader impacts}
    \item[] Question: Does the paper discuss both potential positive societal impacts and negative societal impacts of the work performed?
    \item[] Answer: \textcolor{blue}{[Yes]}
    \item[] Justification: Main paper Section~7 (Discussion, Broader Impact) discusses potential positive impacts (clinical decision support, education, ML research) and negative impacts (risks of direct clinical use without expert oversight; limitations as literature summaries vs.\ peer-reviewed clinical references).

\item {\bf Safeguards}
    \item[] Question: Does the paper describe safeguards that have been put in place for responsible release of data or models that have a high risk for misuse?
    \item[] Answer: \textcolor{blue}{[Yes]}
    \item[] Justification: Dataset release includes prominent disclaimer that ChronoMedKG is a literature-summarized resource and must not be used for patient care without clinical expert oversight. All triples include PMID provenance for verification.

\item {\bf Licenses for existing assets}
    \item[] Question: Are the creators or original owners of assets (e.g., code, data, models), used in the paper, properly credited and are the license and terms of use explicitly mentioned and properly respected?
    \item[] Answer: \textcolor{blue}{[Yes]}
    \item[] Justification: PrimeKG (CC BY 4.0), HPOA (CC BY 4.0), Orphadata (CC BY-NC 3.0), GeneReviews (institutional license), PMC Open Access (various CC licenses) are all used within their terms. Source databases are cited in main paper Section~2 (Related Work).

\item {\bf New assets}
    \item[] Question: Are new assets introduced in the paper well documented and is the documentation provided alongside the assets?
    \item[] Answer: \textcolor{blue}{[Yes]}
    \item[] Justification: ChronoMedKG dataset is documented via this supplementary material including a full Datasheet for Datasets (Appendix~\ref{app:datasheet}), ChronoTQA benchmark is documented with construction methodology (main paper Section~5) and full question-type breakdown (Appendix~\ref{app:tqa-stats}).

\item {\bf Crowdsourcing and research with human subjects}
    \item[] Question: For crowdsourcing experiments and research with human subjects, does the paper include the full text of instructions given to participants and screenshots, if applicable, as well as details about compensation (if any)?
    \item[] Answer: \textcolor{blue}{[N/A]}
    \item[] Justification: No crowdsourcing or human-subject research was conducted. Clinician evaluation (if included) follows standard academic consultation and is described in methods.

\item {\bf Institutional review board (IRB) approvals or equivalent}
    \item[] Question: Does the paper describe potential risks incurred by study participants, whether such risks were disclosed to the subjects, and whether Institutional Review Board (IRB) approvals (or an equivalent approval/review based on the requirements of your country or institution) were obtained?
    \item[] Answer: \textcolor{blue}{[N/A]}
    \item[] Justification: The resource is built from public-domain biomedical literature; no human subjects were recruited. PMC case reports are published clinical cases with patient privacy handled by the original publishing journals.

\item {\bf Declaration of LLM usage}
    \item[] Question: Does the paper describe the usage of LLMs if it is an important, original, or non-standard component of the core methods in this research?
    \item[] Answer: \textcolor{blue}{[Yes]}
    \item[] Justification: LLMs are central to the pipeline. Main paper Section~3 (Construction) describes extraction by DeepSeek V3 and an OpenAI primary (GPT-4.1-nano in early runs, GPT-4o-mini in later runs), with Claude 3 Haiku as a conditional tiebreaker. The three-LLM novelty-judge panel (main paper Section~4.1) uses DeepSeek V3, GPT-4o-mini, and Claude Haiku 4.5. Model versions, parameters (temperature=0), and prompt templates are in Appendix~\ref{app:prompts}.

\end{enumerate}

\newpage

\section{Datasheet for ChronoMedKG}
\label{app:datasheet}

Following the Datasheets for Datasets template \citep{gebru2021datasheets}.

\subsection{Motivation}

\textbf{For what purpose was the dataset created?} ChronoMedKG was created to fill a structural gap in biomedical knowledge graphs: no existing resource captures \emph{when} in the disease course clinical facts are true (onset ages, progression stages, milestone timing). Existing KGs treat all associations as static and timeless, limiting their utility for temporal clinical reasoning tasks.

\textbf{Who created the dataset?} The author team listed on the main paper title page (University of Southern Denmark and University of Hamburg).

\textbf{Who funded the creation of the dataset?} Screen4Care has received funding from the Innovative Medicines Initiative 2 Joint Undertaking (JU) under grant agreement No 101034427. The JU receives support from the European Union’s Horizon 2020 research
and innovation programme and EFPIA. The funders had no role in study design, data collection and analysis, decision to publish, or preparation of the manuscript.
Total direct costs: $\sim$\$2{,}400 (LLM API calls, all-in across construction including retries and failed extractions; \$0.18/disease average) plus researcher time.

\subsection{Composition}

\textbf{What do the instances that comprise the dataset represent?} Each instance is a \emph{temporally-grounded biomedical knowledge triple}: (subject, relation, object) with associated temporal metadata (onset age range, progression stage, milestone, evidence year), evidence provenance (PMID, study type, credibility score), and multi-LLM consensus confidence.

\textbf{How many instances are there in total?} 460{,}497 validated consensus triples derived from 13{,}431 of PrimeKG's 17{,}080 diseases processed by the pipeline (10{,}852 of those diseases retain surviving triples after multi-LLM consensus and Quality Controller filtering); 253{,}094 unique entities (diseases, phenotypes, drugs, genes, anatomical terms, biological processes) across 10 relation types.

\textbf{Does the dataset contain all possible instances or is it a sample?} It is a sample of biomedical knowledge. Coverage is limited by (a) PubMed's indexed literature, (b) the 4-agent pipeline's extraction scope (13{,}431 of 17{,}080 PrimeKG diseases processed), and (c) the consensus filter (only triples with $\geq$2-model agreement retained).

\textbf{What data does each instance consist of?} See main paper Section~3.2. Each triple has: subject, relation, object, temporal metadata (\texttt{onset\_age\_min}, \texttt{onset\_age\_max}, \texttt{progression\_stage}, \texttt{milestone}), evidence (\texttt{source\_ids} list of PMIDs, \texttt{credibility\_score}, \texttt{study\_type}, \texttt{consensus\_confidence}), and quality grade (A = PrimeKG-confirmed, B = novel).

\textbf{Is there a label or target associated with each instance?} Triples are not labeled instances for supervised learning; the resource is a knowledge graph. For downstream tasks (e.g., link prediction), triples serve as observed positive examples in the graph.

\textbf{Is any information missing from individual instances?} NA

\textbf{Are relationships between individual instances made explicit?} Yes: triples share entities, forming a graph. Edges with the same (head, relation) but different (tail) targets represent alternative related entities.

\textbf{Are there recommended data splits?} For link prediction experiments, we use 80/10/10 random splits with random seeds 42, 7, 123 (Appendix~\ref{app:linkpred-details}).

\textbf{Are there any errors, sources of noise, or redundancies in the dataset?} Yes. Error taxonomy (main paper Section~4.3): 7.3\% genuine errors. Additional sources of noise: (a) entity name variants for synonymous concepts (e.g., ``proximal weakness'' vs.\ ``proximal muscle weakness''), (b) onset age ranges from individual papers that may not represent the general disease population, (c) single-triple noise pulling aggregated disease-level ranges.

\textbf{Is the dataset self-contained, or does it link to or otherwise rely on external resources?} Partially self-contained. Each triple links to one or more PubMed PMIDs (external) and entity IDs that align with PrimeKG (external). The full KG is downloadable as a JSONL file; PMID verification requires the PubMed API.

\textbf{Does the dataset contain data that might be considered confidential?} No. All source material is public-domain biomedical literature.

\textbf{Does the dataset contain data that, if viewed directly, might be offensive, insulting, threatening, or might otherwise cause anxiety?} No.

\textbf{Does the dataset identify any subpopulations (e.g., by age, gender)?} No. The dataset describes diseases, not individuals. Age-specific onset ranges apply to disease populations, not to individuals.

\textbf{Is it possible to identify individuals, either directly or indirectly (i.e., in combination with other data) from the dataset?} No.

\textbf{Does the dataset contain data that might be considered sensitive in any way?} No.

\subsection{Collection Process}

\textbf{How was the data associated with each instance acquired?} Via a 4-agent automated pipeline (main paper Section~3) that: (1) profiles each disease against ontologies (MONDO, OMIM, HPO), (2) retrieves PubMed abstracts and PMC full-text articles via NCBI E-utilities, (3) extracts candidate triples using 2--3 LLMs in parallel, (4) applies consensus filtering and PrimeKG schema alignment.

\textbf{What mechanisms or procedures were used to collect the data?} Python scripts using OpenAI, DeepSeek, Anthropic, and Google Gemini APIs. NCBI E-utilities (PubMed, PMC) with registered API key.

\textbf{If the dataset is a sample from a larger set, what was the sampling strategy?} Literature-adaptive tiering: diseases with $\geq$100 PubMed articles received up to 150 documents; diseases with 20--99 articles received all available; diseases with $<$20 articles received all available (exhaustive coverage for rare diseases).

\textbf{Who was involved in the data collection process and how were they compensated?} The authors (primary investigators). No external annotators were used (all labels are from multi-LLM consensus). LLM API costs: $\sim$\$2{,}400 total (paid out of research budget).

\textbf{Over what timeframe was the data collected?} PubMed literature spanning 1960--2026 (median publication year 2015). Extraction pipeline ran February--April 2026.

\textbf{Were any ethical review processes conducted?} Not required: the dataset is built from public-domain literature with no human subjects.

\subsection{Preprocessing/cleaning/labeling}

\textbf{Was any preprocessing/cleaning/labeling of the data done?} Yes. (1) Entity normalization: lowercase, strip parentheticals, slash-split multi-term entities. (2) Consensus filtering: only triples with $\geq$2 LLMs agreeing (fuzzy match $\geq$80\%) pass. (3) Temporal plausibility: age ranges outside [0, 120] filtered. (4) PrimeKG alignment: entity types mapped to PrimeKG's schema.

\textbf{Was the ``raw'' data saved in addition to the preprocessed/cleaned/labeled data?} Yes. 13{,}045{,}687 raw triples (pre-consensus) are retained in per-disease \texttt{raw\_triples.jsonl} files, enabling re-analysis with different consensus thresholds.

\textbf{Is the software used to preprocess/clean/label the instances available?} Yes: full pipeline code at the project repository (Appendix~\ref{app:availability}).

\subsection{Uses}

\textbf{Has the dataset been used for any tasks already?} In this paper: (1) validation against 3 gold standards, (2) ChronoTQA benchmark construction, (3) coverage gap analysis, (4) evidence decay audit, (5) unsupervised trajectory clustering, (6) KG link prediction with temporal features.

\textbf{Is there a repository that links to any or all papers or systems that use the dataset?} Yes: works citing the dataset DOI are tracked automatically via the ``Cited by'' panel on the Zenodo record (\url{https://zenodo.org/records/19697543}), populated through Crossref Event Data and OpenAIRE. A Hugging Face dataset page mirroring the same tracking will be added at camera-ready.

\textbf{What (other) tasks could the dataset be used for?} Temporal differential diagnosis, phenotype progression prediction, stage-specific drug repurposing, clinical decision support systems, educational tools, ML benchmarks for temporal biomedical reasoning, computational phenotyping with temporal features.

\textbf{Is there anything about the composition of the dataset or the way it was collected and preprocessed/cleaned/labeled that might impact future uses?} Yes. (1) ChronoMedKG phenotype entities use LLM-extracted names, not HPO-controlled vocabulary; future users should consider canonicalization (e.g., via SapBERT). (2) The resource should not be used for direct clinical decision-making without expert oversight. (3) 7.3\% of onset ranges are genuinely wrong; aggregate/median-based usage is more robust than single-triple reliance.

\textbf{Are there tasks for which the dataset should not be used?} Direct patient care decisions, diagnostic certainty claims, or any clinical use without peer-reviewed validation and expert oversight.

\subsection{Distribution}

\textbf{Will the dataset be distributed to third parties outside of the entity on behalf of which the dataset was created?} Yes: ChronoMedKG v0.0.1 is already publicly released on Zenodo under CC~BY~4.0 (\url{https://zenodo.org/records/19697543}); a Hugging Face mirror exposing the \texttt{datasets} \texttt{load\_dataset()} API will be added at camera-ready.

\textbf{How will the dataset be distributed?} Zenodo DOI, HuggingFace Datasets, and GitLab repository. Format: JSONL for triples, CSV for tabular summaries, Parquet for large-scale queries.

\textbf{When will the dataset be distributed?} v0.0.1 is already distributed on Zenodo (CC~BY~4.0, public); subsequent versioned releases follow the timeline in \texttt{rai:dataReleaseTimeline} (Croissant metadata).

\textbf{Will the dataset be distributed under a copyright or other intellectual property (IP) license?} Dataset: CC BY 4.0. Code: MIT License.

\textbf{Have any third parties imposed IP-based or other restrictions on the data associated with the instances?} No.

\textbf{Do any export controls or other regulatory restrictions apply to the dataset or to individual instances?} No.

\subsection{Maintenance}

\textbf{Who will be supporting/hosting/maintaining the dataset?} M.~S.~Ahmed (University of Southern Denmark, \texttt{shamim@imada.sdu.dk}) maintains the dataset on Zenodo under the concept DOI \url{https://doi.org/10.5281/zenodo.19697542}, which always resolves to the latest version. New versions and errata are published as fresh Zenodo deposits; a Hugging Face mirror surfacing the same release timeline will be added at camera-ready.

\textbf{How can the owner/curator/manager of the dataset be contacted?} By email at \texttt{shamim@imada.sdu.dk}, or via the GitLab issue tracker at \url{https://gitlab.sdu.dk/screen4care/chronomedkg/-/issues}.

\textbf{Is there an erratum?} The decision log (\texttt{docs/decision\_log.md}) in the public code repository tracks corrections made during development. Version-specific errata for released artefacts are published on Zenodo as new deposits under the concept DOI; the Hugging Face mirror added at camera-ready will surface the same release history.

\textbf{Will the dataset be updated?} Planned updates: (1) entity canonicalisation via SapBERT and \texttt{citation\_count} / \texttt{is\_retracted} credibility-signal backfill (v1.1), (2) activation of the GeneReviews / OMIM Deep tier (v1.2), (3) clinician-validated subset (v2.0). All released versions are preserved as frozen Zenodo DOIs.

\textbf{If the dataset relates to people, are there applicable limits on the retention of the data?} N/A (no personal data).

\textbf{Will older versions of the dataset continue to be supported/hosted/maintained?} Yes: all released versions are preserved on Zenodo with unique DOIs.

\textbf{If others want to extend/augment/build on/contribute to the dataset, is there a mechanism for them to do so?} Yes: GitLab merge requests and issue tracker.

\newpage

\section{Extended Methods}
\label{app:methods}

\subsection{Full Schema Example}
\label{app:schema-example}

A real ChronoMedKG record (Becker muscular dystrophy $\to$ cardiomyopathy, edge \texttt{ef58608a735b}) reproduced verbatim from the released \texttt{validated\_triples.jsonl}, illustrating every field:

\begin{small}
\begin{verbatim}
{"edge_id": "ef58608a735b",
 "source_id": "10311", "source_type": "disease",
 "source_name": "Becker muscular dystrophy",
 "relation": "disease_phenotype_positive",
 "target_id": "1638", "target_type": "phenotype",
 "target_name": "cardiomyopathy",
 "temporal": {"onset_age_min": 20, "onset_age_max": 40,
              "progression_stage": "adult",
              "milestone": "cardiac involvement",
              "temporal_qualifier": null,
              "discovery_date": null, "validity_start": null,
              "validity_end": null, "superseded_by": null,
              "temporal_resolution": "unknown",
              "duration": null, "treatment_start_age": null},
 "evidence": {"tier": 2,
              "source_ids": ["PMID:38224155"],
              "evidence_text": "Cardiac involvement in BMD often
                manifests in the third to fourth decade",
              "study_type": "review",
              "credibility_score": 0.395,
              "consensus_confidence": 1.0,
              "extraction_models": ["claude-haiku"],
              "extraction_method": "tier2_llm_consensus",
              "citation_count": null, "is_retracted": false},
 "conditions": null,
 "extraction_date": "2026-04-03",
 "pipeline_version": "1.0.0",
 "disease_profile_id": "MONDO:10311",
 "quality_grade": "A"}
\end{verbatim}
\end{small}

\noindent Two notes on this v1.0 record. (a)~\texttt{extraction\_models} stores the model whose surviving extraction was kept as the representative row; the canonical record of multi-LLM agreement is \texttt{consensus\_confidence} (1.0 = full agreement across the models that processed the document), with the relationship between the two fields detailed below. (b)~Two of six credibility signals (\texttt{citation\_count}, \texttt{is\_retracted}) are not populated in v1.0 and await v1.1 backfill, as flagged in Limitation~(v) of the main paper.

\begin{figure}[h]
    \centering
    \includegraphics[width=\linewidth]{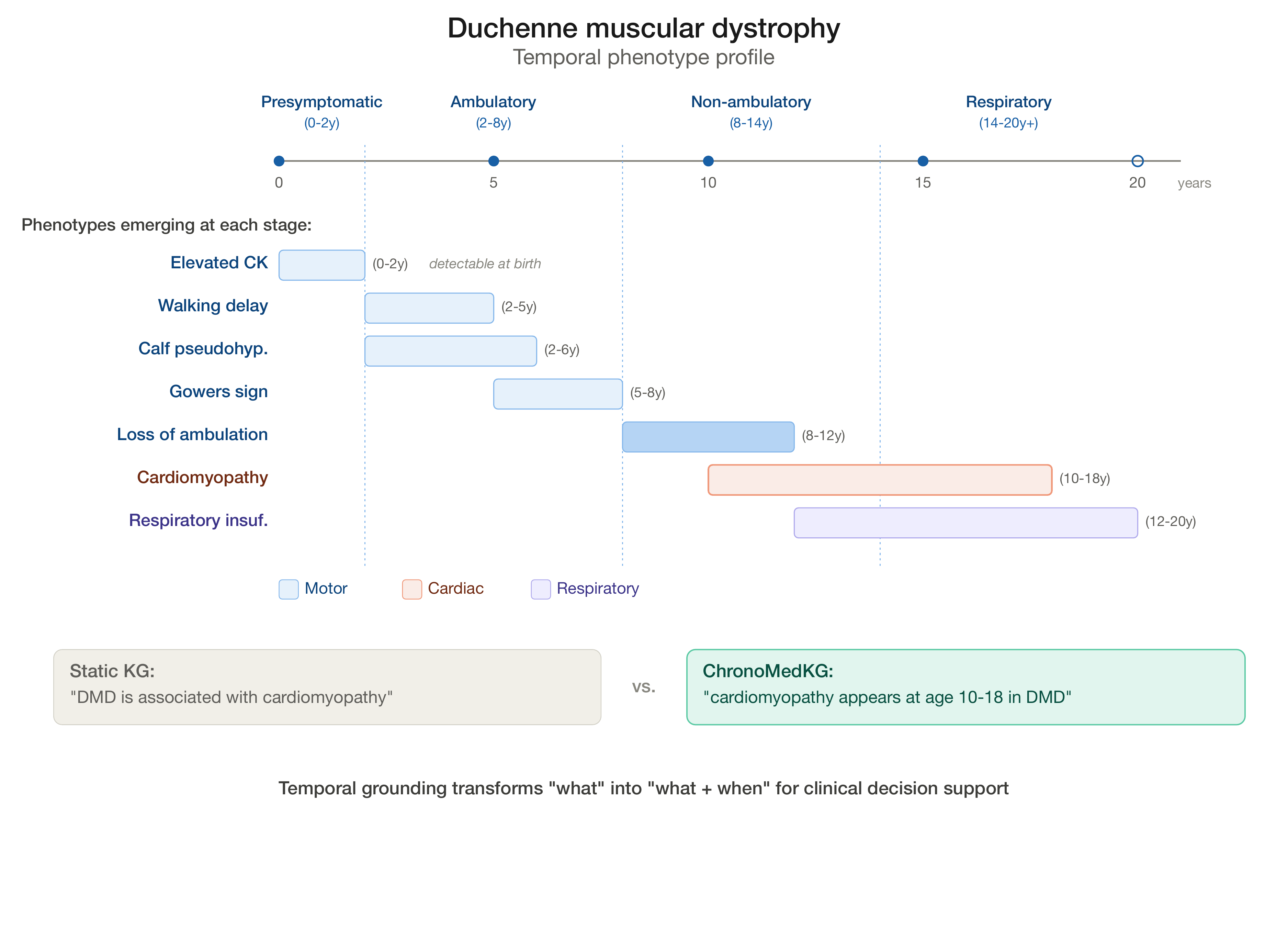}
    \caption{\textbf{Example temporal phenotype profile for Duchenne muscular dystrophy.} Stages (presymptomatic, ambulatory, non-ambulatory, respiratory) partition the age axis; phenotype bars show the onset window for each clinical feature, colour-coded by system (motor, cardiac, respiratory). The static-KG statement ``DMD is associated with cardiomyopathy'' is replaced in ChronoMedKG by a stage-aware statement: ``cardiomyopathy appears at age 10--18 in DMD'', which supports temporal differential diagnosis and age-specific screening decisions that static KGs cannot.}
    \label{fig:dmd_tpp}
\end{figure}

\subsection{4-Agent Pipeline Details}

\paragraph{Agent 1 - Disease Profiler.}
For each disease identifier, queries MONDO, OMIM, and HPO ontologies to generate a YAML configuration containing: disease name, synonyms, differential diagnoses, known genes, key phenotypes, PubMed article count, PMC full-text availability, and coverage flag (rich/moderate/sparse). No LLM calls.

\paragraph{Agent 2 - Evidence Harvester.}
Retrieves abstracts via NCBI E-utilities (\texttt{esearch} + \texttt{efetch}). For diseases with sufficient literature, also retrieves PMC full-text via the PMC Open Access Subset API. Documents are ranked by credibility score (journal tier, citation velocity) and capped per tier: standard tier gets up to 150 documents, light and minimal tiers get all available.

\paragraph{Agent 3 - Knowledge Extractor.}
Each document is sent in parallel to 2--3 LLMs: DeepSeek V3 (primary, low cost), GPT-4.1-nano or GPT-4o-mini (primary), Claude 3 Haiku (conditional tiebreaker, invoked when $\geq$1 primary model returns 0 triples). All LLMs use temperature 0 and 120-second timeout. Extraction prompts are in Appendix~\ref{app:prompts}.

\paragraph{Agent 4 - Quality Controller.}
Applies per-triple validation: (1) non-empty entities, (2) age bounds [0, 120], (3) self-reference prevention, (4) PrimeKG schema alignment (Grade A if confirmed, B if novel), (5) credibility scoring.

\subsection{Multi-LLM Consensus Algorithm}
\label{app:consensus}

\begin{lstlisting}[basicstyle=\small\ttfamily, breaklines=true]
function compute_consensus(per_model_triples, threshold=2):
    # Step 1: collect candidates
    candidates = [(s, r, o, model) for model, triples in per_model_triples
                                   for s, r, o in triples]
    # Step 2: normalize
    normalized = [(normalize(s), canonicalize(r), normalize(o), model)
                  for s, r, o, model in candidates]
    # Step 3: Union-Find clustering by entity similarity
    clusters = UnionFind()
    for i, j in pairs(normalized):
        if relation[i] == relation[j] and
           same_model(i, j) is False and
           fuzz.ratio(s[i], s[j]) >= 80 and
           fuzz.ratio(o[i], o[j]) >= 80:
            clusters.union(i, j)
    # Step 4: filter by consensus threshold
    consensus = []
    for cluster in clusters.groups():
        models_agreeing = set(model for _, _, _, model in cluster)
        if len(models_agreeing) >= threshold:
            representative = max(cluster, key=lambda t: t.confidence)
            representative.confidence = len(models_agreeing) / total_models
            consensus.append(representative)
    return consensus
\end{lstlisting}

Default threshold is 2 (at least 2 distinct LLMs must agree). Fuzzy match threshold of 80\% is based on preliminary calibration: stricter thresholds (90\%) exclude legitimate synonym variants; looser (70\%) admit false positives. The 3.53\% consensus retention rate (460{,}497 of 13.05M raw triples, main paper \S\ref{sec:construction}) reflects per-model duplicate-fact collapse under fuzzy match plus single-model triples filtered by the $\geq$2-model threshold, not signal loss.

\subsection{Six-Signal Credibility Scoring}

For each source document, credibility = weighted combination:
\begin{align*}
\text{credibility} = &\ 0.15 \cdot \text{journal\_tier} + 0.15 \cdot \text{citation\_velocity} \\
+ &\ 0.25 \cdot \text{study\_type\_weight} + 0.15 \cdot \text{replication\_signal} \\
+ &\ 0.15 \cdot \text{retraction\_check} + 0.15 \cdot \text{llm\_consensus}
\end{align*}

Study type weights: meta-analysis 1.0 $>$ guideline 0.95 $>$ RCT 0.9 $>$ database 0.85 $>$ cohort 0.7 $>$ case-control 0.6 $>$ review 0.5 $>$ case-series 0.4 $>$ case-report 0.3 $>$ expert-opinion 0.2 $>$ other 0.1.

\subsection{Link Prediction Experimental Setup}
\label{app:linkpred-details}

\textbf{Data:} 1{,}302 diseases present in both HPOA and ChronoMedKG (disease name normalized: lowercase, common suffixes removed). HPOA edges: 31{,}061 structure / 31{,}077 with onset bins. ChronoMedKG edges: 26{,}184 structure / 28{,}161 with onset bins.

\textbf{Onset bins (aligned between HPOA and ChronoMedKG):} congenital, neonatal, infantile, early\_childhood (ChronoMedKG only), childhood, juvenile, young\_adult, adult, middle\_age (HPOA only), late\_onset.

\textbf{Split:} 80\% train / 10\% validation / 10\% test; one fixed random split per seed (seeds listed below).

\textbf{Model:} TransE (PyKEEN implementation), embedding dimension 100, margin 1.0.

\textbf{Training:} 100 epochs, batch size 1024, Adam optimizer at lr=0.01, negative sampling ratio 1:1.

\textbf{Evaluation:} Hits@1, Hits@3, Hits@10, Mean Reciprocal Rank (MRR). Computed on both head and tail prediction, averaged.

\textbf{Seeds:} 42, 7, 123 (3 independent runs per condition).

\textbf{Statistical testing:} Paired $t$-test on MRR across seeds, treating HPOA-struct $\to$ HPOA-temporal and ChronoMedKG-struct $\to$ ChronoMedKG-temporal as paired observations.

\textbf{Hardware:} Apple M4 (PyTorch MPS backend). Each 100-epoch training takes $\sim$15 seconds; the full 12-run experiment ($2$ sources $\times$ $2$ conditions $\times$ $3$ seeds) completes in $\sim$3 minutes.

\newpage

\section{Extended ChronoTQA Benchmark Statistics}
\label{app:tqa-stats}

\subsection{Biomedical QA Benchmark Landscape}
\label{app:bench-landscape}

ChronoTQA situated alongside the other biomedical QA benchmarks discussed in main paper \S2: ChronoTQA is the only benchmark with age- or stage-conditioned reasoning and dual Tier~1 (external gold standard) + Tier~2 (PMID-traceable KG) grounding.

\begin{table}[h]
\centering
\small
\caption{Biomedical QA benchmarks. ChronoTQA is the only one with age/stage-conditioned reasoning and Tier~1 external + Tier~2 PMID-traceable grounding.}
\label{tab:bench-compare}
\setlength{\tabcolsep}{4pt}
\begin{tabular}{lrllcc}
\toprule
\textbf{Benchmark} & \textbf{N} & \textbf{Format} & \textbf{Domain} & \textbf{Source / Q} & \textbf{Temporal} \\
\midrule
MedQA \citep{medqa} & 12{,}723 & MCQ & USMLE clinical & --- & No \\
PubMedQA \citep{pubmedqa} & 1{,}000\textsuperscript{*} & Yes/No/Maybe & Biomedical literature & Per-abstract & No \\
BioASQ \citep{bioasq} & $\sim$5k & Mixed & Biomedical & Per-snippet & No \\
MMLU-Medical \citep{hendrycks2021mmlu} & $\sim$1k & MCQ & Medical knowledge & --- & No \\
RareBench \citep{rarebench2024} & $\sim$5.4k & Multi-task & Rare-disease Dx & Per-case & No \\
\midrule
\textbf{ChronoTQA} & \textbf{3{,}341} & \textbf{MCQ + free-text} & \textbf{Temporal clinical} & \textbf{Tier 1 ext / Tier 2 PMID} & \textbf{Yes} \\
\bottomrule
\end{tabular}
\\[2pt]
{\footnotesize \textsuperscript{*}PQA-L expert-labeled subset; the artificially-generated PQA-A pool adds $\sim$211k.}
\end{table}

\paragraph{Head-to-head with concurrent agentic KG systems.} Direct comparison of ChronoMedKG against KARMA, MedKGent, or AutoBioKG on ChronoTQA is not feasible: none of the three releases exposes a temporal-query interface, and each encodes a different temporal or contextual axis (main paper \S2). Reproducing the systems against ChronoTQA's temporal demands would require re-implementing their pipelines, which is out of scope for the v1 release.

\subsection{Question Difficulty Distribution}

Each ChronoTQA question is labeled with a difficulty level based on automated heuristics:

\begin{itemize}
    \item \textbf{Easy:} Common diseases, standard onset ranges (e.g., DMD age 3--5 years)
    \item \textbf{Medium:} Moderate-frequency diseases, some variability in onset literature
    \item \textbf{Hard:} Rare diseases, limited literature, or ambiguous onset ranges
\end{itemize}

\begin{table}[h]
\centering
\caption{Difficulty distribution across ChronoTQA question types. Difficulty labels are assigned per question type based on the reasoning demand: exact onset-window and ordering subtasks are \emph{hard} (require precise age ranges or sequences), coarse comparisons are \emph{medium}, and single-fact static lookups are \emph{easy}.}
\begin{tabular}{lrrrr}
\toprule
\textbf{Type} & \textbf{Easy} & \textbf{Medium} & \textbf{Hard} & \textbf{Total} \\
\midrule
Temporal window & 0 & 800 & 0 & 800 \\
Temporal differential Dx & 0 & 0 & 687 & 687 \\
Cross-disease comparison & 0 & 600 & 0 & 600 \\
Phenotype ordering & 0 & 0 & 395 & 395 \\
Stage-conditional & 0 & 0 & 200 & 200 \\
Phenopackets onset & 0 & 0 & 147 & 147 \\
Negative temporal MCQ & 0 & 0 & 12 & 12 \\
Static control drug & 250 & 0 & 0 & 250 \\
Static control gene & 250 & 0 & 0 & 250 \\
\midrule
\textbf{Total} & \textbf{500} & \textbf{1{,}400} & \textbf{1{,}441} & \textbf{3{,}341} \\
\bottomrule
\end{tabular}
\end{table}

\subsection{Disease Coverage in ChronoTQA}

Of the 13{,}431 PrimeKG diseases the pipeline processed, 2{,}315 appear in at least one ChronoTQA question. Coverage per disease averages 1.44 questions.

\subsection{Example Questions per Type}

\textbf{Temporal window:}
``Is age 8 years within the typical onset window for Kleefstra Syndrome Due To 9q34 Microdeletion?''
Gold: No (0--2 years).

\textbf{Temporal differential Dx:}
``A patient presents with symptoms during the prenatal period. Based on typical age of onset, which of the following diseases is most consistent: (A) Acute Zonal Occult Outer Retinopathy, (B) Lipodystrophy Due To Peptidic Growth Factors Deficiency, (C) Isolated Congenital Microcephaly, (D) Trichofolliculoma?''
Gold: C.

\textbf{Cross-disease comparison:}
``Which disease typically has an earlier age of onset: Maffucci Syndrome or Congenital Hydrocephalus?''
Gold: Congenital Hydrocephalus.

\textbf{Phenopackets onset:}
``At what age does `proptosis' typically present in Chitayat Syndrome? (Based on patient case data)''
Gold: 0.0--2.7 years (from 2 cases).

\textbf{Stage-conditional:}
``What phenotypes are characteristic of the non-ambulatory stage of Duchenne Muscular Dystrophy?''
Gold: Loss of independent ambulation, joint contractures, Gowers sign.

\textbf{Phenotype ordering:}
``Rank the following clinical milestones for Duchenne Muscular Dystrophy by typical age of occurrence: diagnosis, loss of ambulation, cardiomyopathy onset.''
Gold: diagnosis $\rightarrow$ loss of ambulation $\rightarrow$ cardiomyopathy.

\newpage

\section{Error Analysis: Concrete Examples}
\label{app:errors}

Below are representative examples from each error taxonomy category (main paper Table~5). Examples are drawn from the 2{,}563 ChronoMedKG--Orphadata matched diseases.

\subsection{Correct (contained): 50.1\%}
Example: Duchenne muscular dystrophy. ChronoMedKG median range: 2--5y. Orphadata: 1--5y. Within gold standard range.

\subsection{Adjacent stage: 15.6\%}
Example: Disease X. ChronoMedKG median: 2--12y (childhood). Orphadata: 5--15y (juvenile). Ranges overlap at 5--12y but ChronoMedKG extends slightly earlier and gold slightly later. Clinically borderline, not a real error.

\subsection{Granularity mismatch: 13.8\%}
Example: Duchenne muscular dystrophy. Orphadata gives one range (1--5y) for the disease. ChronoMedKG gives per-phenotype ranges: walking delay 2--5y, cardiomyopathy 10--18y. Aggregated ChronoMedKG range (2--18y) does not fit Orphadata's 1--5y, but neither is wrong: ChronoMedKG just tracks more detail.

\subsection{ChronoMedKG wider but overlaps: 6.7\%}
Example: Marfan syndrome. Orphadata: childhood (5--12y). ChronoMedKG: 0--40y because ChronoMedKG captures congenital features (aortic dilation detected prenatally) and adult features (progressive aortic root dilation). Core childhood period still overlaps.

\subsection{Single-triple noise: 5.7\%}
Example: Disease Y. ChronoMedKG has 50 triples about this disease; 49 have onset in 2--8y, but one outlier triple has onset at 60y (possibly misattributed from a comorbid condition). Median aggregation still correct; min/max includes the noise.

\subsection{Genuinely wrong: 7.3\%}
Example: Disease Z. ChronoMedKG median range: 30--60y (adult onset). Orphadata: 0--1y (neonatal). No overlap, $>$10-year gap. Likely due to extraction from papers where patient age at report was conflated with disease onset age.

\newpage

\section{PMC Clinical Case Studies (All 31 Cases)}
\label{app:pmc-cases}

\begin{figure}[h]
    \centering
    \includegraphics[width=\linewidth]{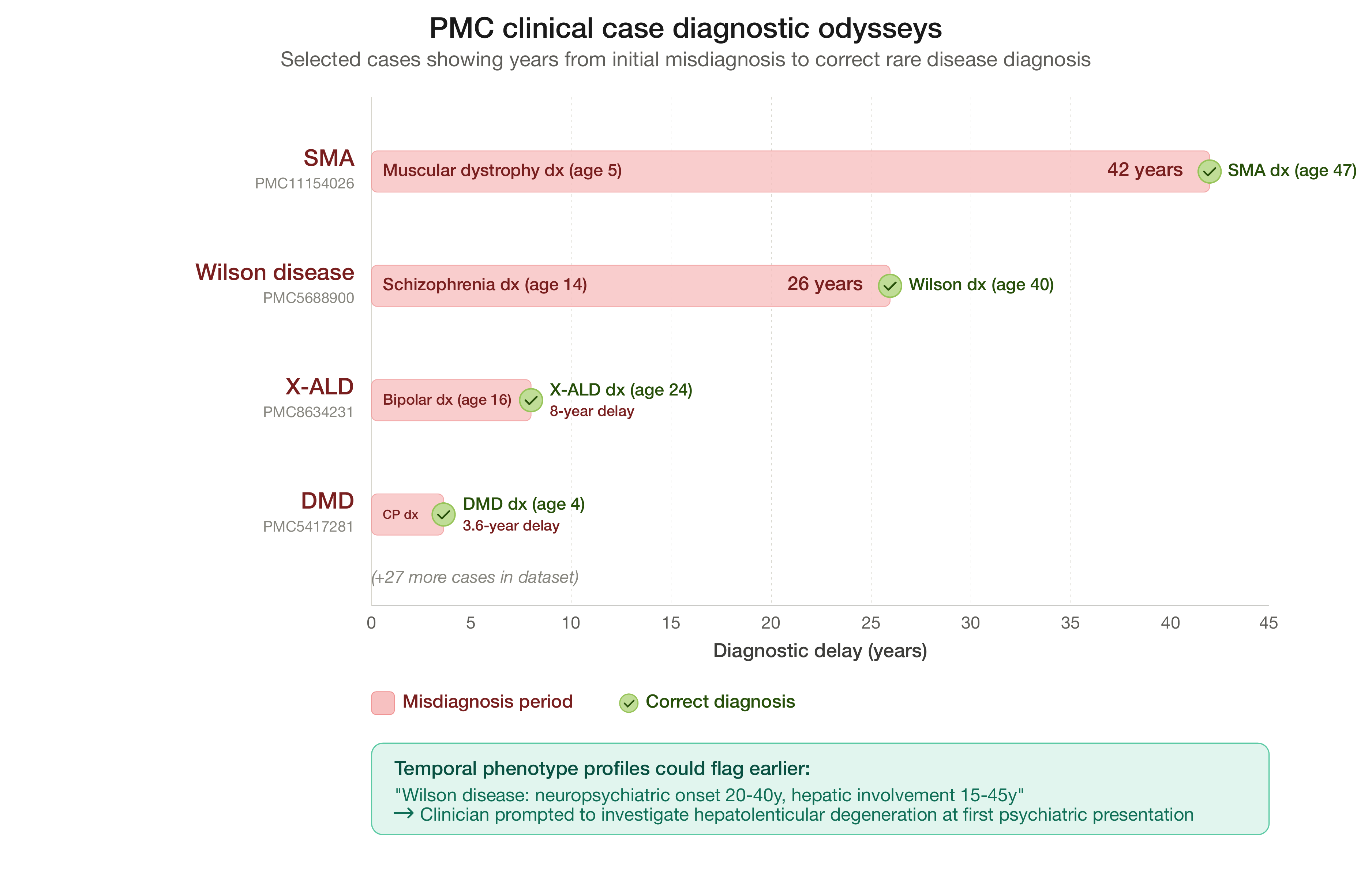}
    \caption{Diagnostic odysseys in four representative PMC cases. Pink bars show the misdiagnosis period (from initial presentation to correct diagnosis); green check marks the correct diagnosis. Delays range from 3.6 to 42 years across 31 cases (median 14y; full list in Table~\ref{tab:pmc-all}). The green callout illustrates how a ChronoMedKG phenotype profile could prompt earlier investigation, using Wilson disease as the example.}
    \label{fig:pmc-timelines}
\end{figure}

\begin{longtable}{llrll}
\caption{All 31 PMC open-access clinical case reports used as qualitative case studies. Diagnostic delay is the number of years between initial symptom onset and correct diagnosis. All cases are verifiable at \texttt{https://pmc.ncbi.nlm.nih.gov/articles/<PMC\_ID>/}. \label{tab:pmc-all}} \\
\toprule
\textbf{PMC ID} & \textbf{Correct Diagnosis} & \textbf{Delay} & \textbf{Misdiagnosis} & \textbf{Category} \\
\midrule
\endfirsthead
\multicolumn{5}{c}{\tablename\ \thetable\ -- continued} \\
\toprule
\textbf{PMC ID} & \textbf{Correct Diagnosis} & \textbf{Delay} & \textbf{Misdiagnosis} & \textbf{Category} \\
\midrule
\endhead
\bottomrule
\endfoot

PMC5417281 & Duchenne muscular dystrophy & 3.6y & Cerebral palsy + epilepsy & Neuromuscular \\
PMC6243516 & LGMD2A (calpainopathy) & 8y & Duchenne muscular dystrophy & Neuromuscular \\
PMC5526785 & Oculopharyngeal MD & 8y & Myasthenia gravis & Neuromuscular \\
PMC5787973 & Duchenne muscular dystrophy & --- & None (reference) & Neuromuscular \\
PMC3454578 & Becker MD w/ cardiomyopathy & --- & None (reference) & Neuromuscular \\
PMC8597609 & Myhre syndrome (SMAD4) & 1.5y & Marfan syndrome & Connective tissue \\
PMC4639334 & Late-onset Pompe disease & 14y & Muscular dystrophy & Lysosomal \\
PMC11154026 & Spinal muscular atrophy & 42y & Muscular dystrophy & Neuromuscular \\
PMC9468738 & Vascular EDS & 22.5y & Hemophilia (negative workup) & Connective tissue \\
PMC5688900 & Wilson disease & 26y & Schizophrenia & Neurogenetic \\
PMC12131641 & Late-onset Fabry disease & 16y & Type 2 diabetic nephropathy & Lysosomal \\
PMC3756019 & Fabry disease (heterozygous) & 5y & Multiple sclerosis & Lysosomal \\
PMC10519872 & Late-onset MELAS & 2y & Coronary artery disease & Mitochondrial \\
PMC9549677 & MELAS syndrome & 4y & Autoimmune encephalitis & Mitochondrial \\
PMC11938598 & Adult-onset Leigh & 19.5y & Hypertensive encephalopathy & Mitochondrial \\
PMC3669838 & MPS IIIB (Sanfilippo) & 21y & Attention deficit disorder & Lysosomal \\
PMC12831942 & Alstrom syndrome & 24y & None (classical) & Ciliopathy \\
PMC10619069 & X-ALD (cerebral) & 20y & Tumefactive MS & Peroxisomal \\
PMC8634231 & X-ALD (adult cerebral) & 8y & Bipolar affective disorder & Peroxisomal \\
PMC10319965 & Bardet-Biedl syndrome & 14y & None (heterogeneous) & Ciliopathy \\
PMC11196773 & Bardet-Biedl syndrome & 7y & None (heterogeneous) & Ciliopathy \\
PMC11468024 & Long QT syndrome type 2 & 1y & Epilepsy & Channelopathy \\
PMC8418736 & Long QT syndrome type 2 & 22y & Refractory epilepsy & Channelopathy \\
PMC7132987 & Danon disease (LAMP2) & 2y & Sarcomeric HCM & Cardiomyopathy \\
PMC11583564 & Gaucher disease type 1 & 34y & Chronic asthma & Lysosomal \\
PMC11346882 & Gaucher disease & 20y & Idiopathic thrombocytopenia & Lysosomal \\
PMC5994559 & Chronic granulomatous disease & 36y & Tuberculosis / vasculitis & Immunodeficiency \\
PMC3326000 & Chronic granulomatous disease & 50y & Fungal allergy / ABPA & Immunodeficiency \\
PMC11101243 & Common variable immunodefici.\ & 9y & Immune thrombocytopenia & Immunodeficiency \\
PMC3535999 & Osteogenesis imperfecta I & 1.33y & Non-accidental injury & Skeletal dysplasia \\
PMC8047228 & Wiskott-Aldrich syndrome & 3.9y & Pulmonary tuberculosis & Immunodeficiency \\
\end{longtable}

\textbf{Summary:} 26/31 cases involved misdiagnosis. Diagnostic delays range 1--50 years (median 14y). 159 total phenotype-timepoint annotations across 11 disease categories.

\newpage

\section{LLM Extraction Prompts}
\label{app:prompts}

\subsection{Primary Extraction Prompt}

The following prompt is sent to each LLM per document. Variables \texttt{\{disease\_name\}}, \texttt{\{known\_genes\}}, etc.\ are substituted per disease. The PMID and document publication year are \emph{not} part of what the LLM returns: they are attached to every extracted triple by the orchestrator, copied from the source document's NCBI E-utilities metadata. This ensures that PMID provenance is deterministic and cannot be hallucinated by the LLM. Reviewers can verify any quoted \texttt{evidence\_text} against the PMID-linked article directly.

\begin{verbatim}
You are a temporal biomedical knowledge extraction system. Your PRIMARY
task is to extract relationships WITH TEMPORAL GROUNDING from the text
about {disease_name}.

CRITICAL: Every relationship you extract MUST include temporal information
when available.

Extraction priorities (highest to lowest):
1. TEMPORAL FACTS: onset ages, disease milestones, progression timelines,
   treatment timing, discovery dates
2. EVIDENCE-DATED FACTS: relationships anchored by publication year
3. CONDITIONAL FACTS: relationships that depend on age, stage, genetic
   subtype
4. STATIC FACTS: general relationships without temporal context

Output format (JSON):
{
  "triples": [
    {
      "subject": "entity name",
      "subject_type": "disease|gene/protein|drug|phenotype|anatomy|...",
      "relation": "disease_protein|indication|disease_phenotype_positive|...",
      "object": "entity name",
      "object_type": "same vocabulary as subject_type",
      "confidence": "high|medium|low",
      "evidence_text": "exact quote from source (max 200 chars)",
      "temporal_context": {
        "onset_age_min": 3.0,
        "onset_age_max": 12.0,
        "progression_stage": "ambulatory",
        "milestone": "loss of ambulation",
        "discovery_year": 2015,
        "temporal_qualifier": "by age 12"
      },
      "conditions": {
        "age_group": "pediatric",
        "genetic_subtype": "exon deletion"
      }
    }
  ]
}

Disease context:
- Name: {disease_name}
- Category: {disease_category}
- Inheritance: {inheritance_pattern}
- Known genes: {known_genes}
- Key phenotypes: {known_phenotypes}
- Differential diagnoses: {differential_diseases}

Source text:
{text}

Extract ALL temporally-grounded relationships. Return valid JSON only.
\end{verbatim}

\subsection{Temporal Re-extraction Prompt (Second Pass)}

For documents that yielded few temporal triples on the first pass, a second
focused prompt is used:

\begin{verbatim}
SECOND PASS — TEMPORAL ONLY: Find temporal information that was missed.
Extract ONLY relationships with temporal grounding: ages, stages, durations,
progression, milestones. Skip any fact without temporal content.

Output format (JSON): same as primary extraction, but 'temporal_context'
field is REQUIRED (non-null).
\end{verbatim}

\newpage

\section{Extended Evaluation Experiments}
\label{app:eval}

\subsection{RAG Experiment on External Gold-Standard Questions (Tier 1)}
\label{app:rag}

\paragraph{Three-LLM calibrated-scoring RAG leaderboard (147 Phenopackets questions).} We tested RAG performance on the 147 Phenopackets-grounded onset questions in ChronoTQA across three LLMs (Claude~3 Haiku, DeepSeek~V3, GPT-4o-mini) under four retrieval conditions (NR, PrimeKG, HPOA, ChronoMedKG). Answers were scored with a calibrated rubric (overlap tolerance $= \max(0.5,\, 0.5 \cdot w_{\text{gold}})$, capped at $\pm$2y, with category-keyword matching admitted only when keyword width is compatible with the gold range), which corrected two bugs in an earlier scorer that had inflated aggregate accuracy by $\sim$24\%.

\begin{table}[h]
\centering
\small
\caption{RAG leaderboard on 147 Phenopackets-grounded onset questions (calibrated scoring, $n{=}147$ per cell). Pooled accuracy improvements are bounded by high no-retrieval ceilings; the stronger result is long-tail rescue (main paper Table~9).}
\label{tab:rag-leaderboard}
\begin{tabular}{lrrrrr}
\toprule
\textbf{Model} & \textbf{NR} & \textbf{PrimeKG} & \textbf{HPOA} & \textbf{ChronoMedKG} & \textbf{ChronoMedKG$-$NR} \\
\midrule
Claude~3 Haiku & 76.2\% & 68.7\% & 75.5\% & \textbf{81.6\%} & $+$5.4 \\
DeepSeek~V3    & 86.4\% & 81.6\% & 83.7\% & 85.0\%           & $-$1.4 \\
GPT-4o-mini    & 76.9\% & 81.6\% & 81.0\% & 79.6\%           & $+$2.7 \\
\midrule
\textbf{Pooled} ($n{=}441$) & 79.8\% & 77.3\% & 80.1\% & \textbf{82.1\%} & $+$2.3 \\
\bottomrule
\end{tabular}
\end{table}

Claude~3 Haiku is the only model where the per-model ChronoMedKG-vs-PrimeKG gain reaches significance (McNemar exact $p{=}0.008$, $+$12.9pp); pooled, ChronoMedKG-vs-PrimeKG trends positive but is not significant ($p{=}0.06$). PrimeKG-RAG \emph{hurts} accuracy for two of three models, consistent with the observation that retrieving non-temporal associations distracts from temporally-framed questions.

\paragraph{Training-contamination ceiling.} Three LLMs score 76--86\% on Phenopackets-sourced questions with \emph{no retrieval}, a ceiling consistent with the GA4GH Phenopackets corpus having been public since 2019 and likely represented in LLM pretraining. Pooled RAG gains are upper-bounded: there is little headroom for any retrieval source to improve an LLM that already answers 80\%+ from parametric memory. The better test under this ceiling is the long-tail rescue measure (main paper Table~9). On the questions each model cannot answer from pretraining, ChronoMedKG improves outcomes by 47--65\% across three LLM families. We therefore frame ChronoMedKG's RAG contribution as selective: it delivers the largest gain where static KGs and coarse-onset resources fail, not as a pooled-accuracy claim against the ceiling set by what these models already know.

The underlying capability gap this RAG experiment targets is quantified independently in \S\ref{app:llm_gap}: across four frontier LLMs (ChatGPT, Gemini, Claude, DeepSeek) on 120 ChronoTQA items posed without retrieval, temporal accuracy trails static accuracy by a mean $+$30.1\,pp, and free-text Phenopackets onset questions collapse to a 4.5\% mean across models, the exact subtask ChronoMedKG is designed to serve.

\subsection{Parametric-Knowledge Gap Across Four LLMs}
\label{app:llm_gap}

The headline static-vs-temporal gap (main paper Table~7 in Section~5) was measured by sampling 120 questions from ChronoTQA (stratified across temporal-window, temporal differential diagnosis, cross-disease onset comparison, Phenopackets-grounded onset, and static controls) and posing them without retrieval to four LLMs via their web chat interfaces (no API calls, no system prompts beyond the benchmark's formatting instructions): OpenAI GPT-4o-mini (ChatGPT), Google Gemini, Anthropic Claude, and DeepSeek-V3. Answers were scored against the gold label using exact match for multiple-choice and Yes/No items and bidirectional substring match for free-text disease names and onset bins.

The pattern is especially stark on free-text Phenopackets onset questions (``At what age does phenotype $X$ typically present in disease $Y$?''): the four models score 0\%, 12\%, 0\%, and 6\% (mean 4.5\%), against an 80\%+ static-control ceiling (main paper Table~7). This is precisely the task where parametric knowledge is insufficient and where ChronoMedKG provides directly queryable answers with PMID-grounded provenance. Full per-model breakdowns and answer logs are released in \texttt{data/benchmark/chat\_eval\_summary.json}.

\subsection{LLM-as-Judge Calibration}
\label{app:judge}

A DeepSeek V3 judge tested on 100 diseases with known ground truth (50 where ChronoMedKG is objectively correct, 50 where ChronoMedKG is objectively wrong) produces: 77.1\% ``supported'' on correct cases (22.9\% false negative rate) and 46.8\% ``supported'' on wrong cases. The single-judge baseline is too lenient on errors and too strict on correct answers to be a primary validation tool; the three-LLM panel used for the novel-coverage audit (\S\ref{app:novelty-detail}) reduces both directions of bias by requiring verifiable-majority agreement, and the 87.9\% reported in main paper Section~4.1 is taken only after that filter. The 92.7\% Orphadata accuracy uses direct gold-standard comparison rather than LLM judging.

\subsection{Error Taxonomy Visualisation}
\label{app:error-fig}

\begin{figure}[h]
    \centering
    \includegraphics[width=0.95\linewidth]{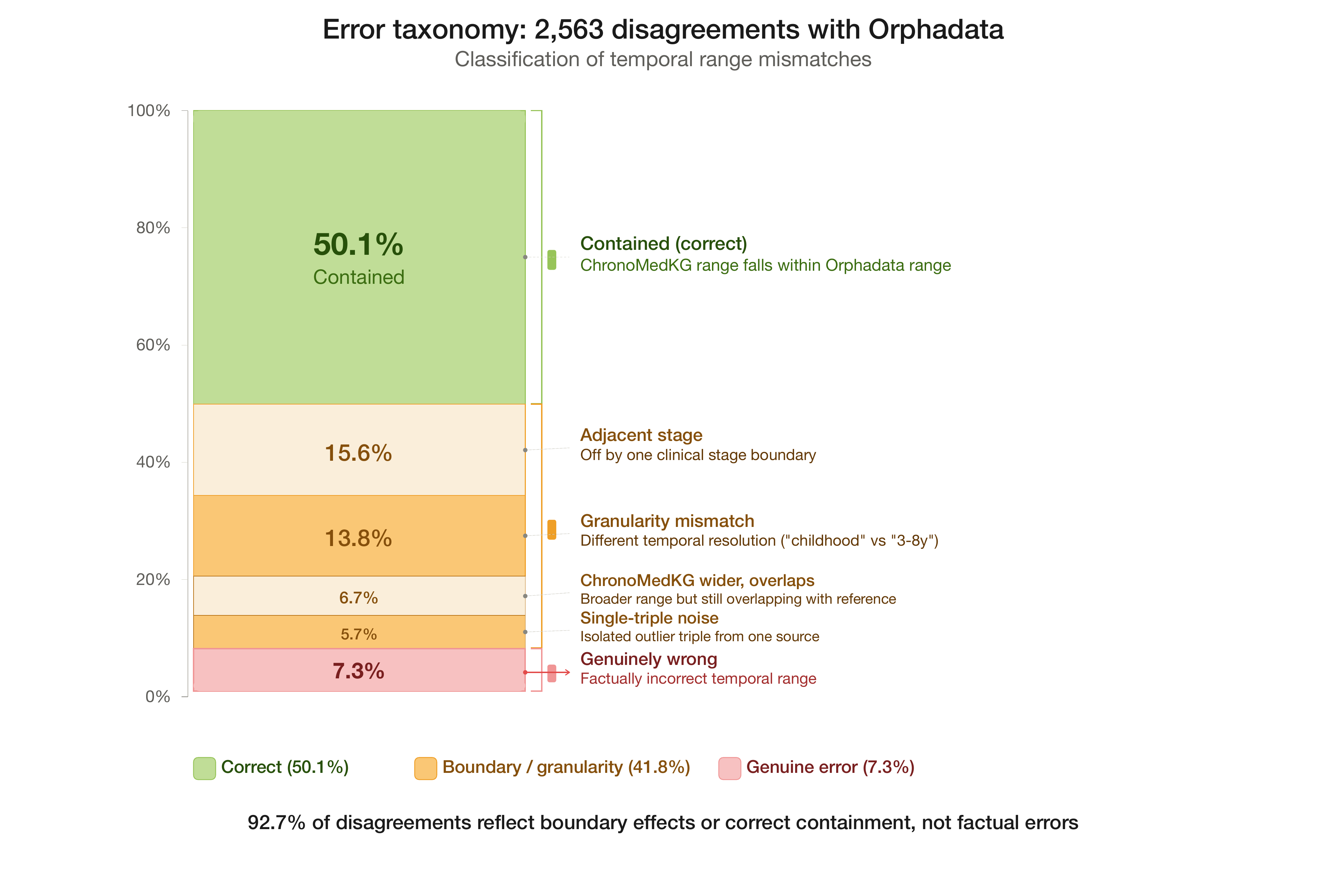}
    \caption{\textbf{Classification of 2{,}563 ChronoMedKG--Orphadata onset-range disagreements.} Only 7.3\% are factually incorrect temporal ranges (no overlap, $>$10y gap from gold). 50.1\% are strictly contained within the gold range, and a further 41.8\% reflect boundary effects or granularity mismatches (e.g., ChronoMedKG's ``3--8y'' vs.\ Orphadata's ``childhood''), not extraction errors. Percentages sum to 99.2\% due to rounding.}
    \label{fig:error_taxonomy}
\end{figure}

\subsection{Novel-Coverage Verification: Full Protocol, Error Taxonomy, Scope}
\label{app:novelty-detail}

\paragraph{Full sampling and protocol.} We drew $n{=}100$ diseases from the 6{,}250 novel-coverage set (seed$=$42), stratified by literature tier (Standard/Light/Minimal) and onset bucket (prenatal, infantile, childhood, adolescent, adult, late-adult). For each disease we selected the single triple whose evidence text contained the most informative timing keyword (longest keyword-bearing span), producing one (claim, evidence) pair per disease. Three judges (DeepSeek~V3, GPT-4o-mini, Claude Haiku~4.5, all temperature~0.0) independently rated each pair under a citation-aware chain-of-thought prompt that required, in order: (1) quote the timing clause from the evidence verbatim, (2) translate the clause to a numeric age range using a fixed clinical-era lookup (prenatal 0--0 y, infancy 0--1, early childhood 1--5, childhood 1--11, adolescence 10--18, adulthood 18--65, older adulthood $\geq$65), (3) compare per triple, not per disease, against the claimed range, and (4) return one of {\scshape supported} / {\scshape partially\_supported} / {\scshape not\_supported} / {\scshape unverifiable}. Per-triple comparison corrects an aggregation bug in a prior single-judge audit that compared the disease-level aggregated median to one triple's evidence.

\paragraph{Full error taxonomy.} 
Every one of the 11 NOT\_SUPPORTED cases failed in the same way. The extractor picked up a qualitative phrase like "elderly onset" or "mid-trimester" and wrote it to the temporal\_qualifier field, but never translated it into the numeric onset\_age\_min/onset\_age\_max fields that the novel-coverage count actually uses. So the evidence text was fine - the pipeline just dropped the numeric range on the floor. This is a normalisation bug downstream of extraction, not an LLM hallucination.
A simple fallback mapping from qualifier strings to clinical-era ranges would reclassify 10 of 11 failures; the one remaining case (MONDO:402) has neither a qualifier nor a numeric range and represents genuine extraction noise (1/100 = 1\%). The 6 three-way-split cases are dominated by disagreement over whether broad qualitative phrases (``adult'', ``late-onset'') sufficiently support narrower numeric claims, which is a semantic question about claim width rather than evidence presence.

\paragraph{Scope of this audit.} The 87.9\% figure applies to the novel-coverage subset only; it is not a re-estimate of the 92.7\% Orphadata accuracy (different denominator, different evidence). Scaling the audit to all 6{,}250 novel-coverage diseases, and to the full 13{,}431-disease resource, is open work; the automation built here (\texttt{scripts/llm\_judge\_novelty\_v2\_multi.py}, 3 providers async, $\sim$\$2 per 100 triples) is released with the resource to let reviewers and downstream users replicate or extend the audit on their own samples.

\subsection{Clinician Validation via HEG-TKG: Full Likert Breakdown}
\label{app:clinician}

The extraction methodology that ChronoMedKG scales was evaluated by a three-clinician panel in HEG-TKG~\citep{ahmed2026hegtkg} on six rare neuromuscular diseases (three pairs: DMD/BMD, MG/LEMS, CIDP/GBS) that are also covered in ChronoMedKG. Each clinician independently rated single system-generated outputs (either vanilla GPT-4.1 or HEG-TKG, blinded to system identity, shuffled with a deterministic seed) on five Likert dimensions: Verifiability (D1), Actionability (D2), Temporal Precision (D3), Non-Expert Safety (D4), and Clinical Completeness (D5). Statistical inference uses Mann--Whitney~U with Benjamini--Hochberg correction across five comparisons and bootstrap 95\% confidence intervals from 10{,}000 resamples.

\begin{table}[h]
\centering
\small
\caption{Clinician ratings of HEG-TKG vs.\ vanilla GPT-4.1 on six shared diseases. C1 = senior neurologist (36-case reduced pack: 18 vanilla, 18 HEG-TKG); C2 = neurologist (72-case full pack: 36 per arm). Scores on 5-point Likert scale; statistics as described in the preceding paragraph.}
\label{tab:clinician}
\setlength{\tabcolsep}{5pt}
\begin{tabular}{llccc}
\toprule
\textbf{Rater} & \textbf{Dimension} & \textbf{Vanilla} & \textbf{HEG-TKG} & \textbf{$\Delta$ (Cohen's $d$, $p_{\mathrm{BH}}$)} \\
\midrule
C1 (senior neurologist) & D1 Verifiability & 2.36 & 4.00 & $+$1.64 ($d{=}1.81$, $p{<}0.001$) \\
C1                       & D2 Actionability & 2.75 & 3.61 & $+$0.86 ($d{=}1.00$, $p{=}0.009$) \\
C1                       & D3 Temporal      & 3.78 & 4.17 & $+$0.39 ($d{=}0.45$, $p{=}0.17$) \\
C1                       & D4 Safety        & 3.17 & 3.61 & $+$0.44 ($d{=}0.57$, $p{=}0.12$) \\
C1                       & D5 Completeness  & 3.25 & 3.47 & $+$0.22 ($d{=}0.35$, $p{=}0.17$) \\
\midrule
C2 (neurologist)         & D1--D5           & \multicolumn{3}{l}{No significant effect on any dimension (all $p_{\mathrm{BH}}>0.79$, $|d|<0.31$)} \\
\bottomrule
\end{tabular}
\end{table}

The senior neurologist (C1) found HEG-TKG markedly more verifiable ($+$1.64/5, $d{=}1.81$, $p{<}0.001$) and more actionable ($+$0.86/5, $d{=}1.00$, $p{=}0.009$) than the vanilla-GPT-4.1 baseline; trends were positive on the remaining three dimensions but did not reach significance after BH correction. The second neurologist (C2) found no significant differences on any dimension (all $p_{\mathrm{BH}}>0.79$, $|d|<0.31$); qualitative notes reported in HEG-TKG attribute this to an evaluation philosophy that treats citations of consensus knowledge as redundant rather than useful (e.g., penalising HEG-TKG for citing standard-of-care treatments with PMIDs). This rater-by-rater divergence is itself the case for the multi-rater clinician-validated subset queued for v2.0 (Limitation~(ii) in main paper Section~7).

\paragraph{Follow-up independent replication on CIDP/GBS.} Subsequent to the 2-clinician panel above, HEG-TKG co-author J.~Zschüntzsch (senior neurologist, 20~years' experience, University Medical Center Göttingen) performed an independent blinded evaluation of the same 12-case CIDP/GBS reduced pack that C1 scored. Following the same protocol (single output per case, blinded to system, shuffled), she rated 11 of 12 cases on all five dimensions (one case left unrated). Her per-dimension stratified means are reported in Table~\ref{tab:clinician-followup}, with C1's matched CIDP/GBS subset for direct comparison. Direction of effect replicates across raters on D1--D4: both raters find HEG-TKG higher than vanilla GPT-4.1 on Verifiability ($\Delta{=}{+}1.30$ vs.\ C1's $\Delta{=}{+}1.58$ on the same 12 cases), with consistent positive deltas on Actionability, Temporal Precision, and Safety. D5 Completeness diverges in sign between raters (Jana $\Delta{=}{+}1.00$, C1 $\Delta{=}{-}0.08$), reflecting small-sample variance. Sample sizes are too small for per-rater significance testing at this subset ($n{=}5$--$6$ per arm); the finding is a \emph{direction-of-effect replication}, not an independent significance claim.

\begin{table}[h]
\centering
\small
\caption{Follow-up independent blinded evaluation by HEG-TKG co-author J.~Zschüntzsch (UMG G\"ottingen) on the same 12-case CIDP/GBS reduced pack that C1 scored. $n{=}6$ vanilla, $n{=}5$ HEG-TKG for Jana (one case unrated); $n{=}6$ per arm for C1. Values are per-dimension Likert means (1--5 scale).}
\label{tab:clinician-followup}
\setlength{\tabcolsep}{5pt}
\begin{tabular}{lccc|ccc}
\toprule
 & \multicolumn{3}{c|}{\textbf{Jana (follow-up)}} & \multicolumn{3}{c}{\textbf{C1 (original, CIDP/GBS subset)}} \\
\textbf{Dimension} & \textbf{Vanilla} & \textbf{HEG-TKG} & \textbf{$\Delta$} & \textbf{Vanilla} & \textbf{HEG-TKG} & \textbf{$\Delta$} \\
\midrule
D1 Verifiability  & 2.50 & 3.80 & $+$1.30 & 2.08 & 3.67 & $+$1.58 \\
D2 Actionability  & 3.00 & 3.80 & $+$0.80 & 2.58 & 3.50 & $+$0.92 \\
D3 Temporal       & 3.83 & 4.40 & $+$0.57 & 4.00 & 4.50 & $+$0.50 \\
D4 Safety         & 2.83 & 3.40 & $+$0.57 & 3.00 & 3.67 & $+$0.67 \\
D5 Completeness   & 3.00 & 4.00 & $+$1.00 & 3.42 & 3.33 & $-$0.08 \\
\bottomrule
\end{tabular}
\end{table}

\paragraph{Scope of clinician evidence.} This validation applies to the HEG-TKG pipeline on six diseases; ChronoMedKG has not been clinician-validated at its 13{,}431-disease scope. The panel result is an anchor for the underlying methodology, not a claim about the resource as a whole. Clinician rating of novel-disease triples, in particular the 6{,}250 diseases where ChronoMedKG provides onset data absent from any gold standard, is the critical remaining validation step and is in progress.

\subsection{Disease Trajectory Clustering: Archetype Table and Sensitivity}
\label{app:clustering}

We clustered 8{,}935 diseases with onset data by temporal signature (median onset, onset spread, number of stages, milestone density, fraction of triples with onset). K-means identified four clinically-coherent archetypes (silhouette 0.362, best of $k\in\{4..8\}$):

\begin{table}[h]
\centering
\small
\caption{Disease trajectory archetypes discovered by unsupervised clustering. \emph{Median onset} is the per-disease median of the earliest-onset triple; \emph{median spread} is the per-disease maximum minus minimum onset across all phenotype triples for that disease. The ``broad-onset progressive'' archetype has a childhood-centred median but a wide spread because these diseases accumulate phenotypes spanning the full life course (e.g., muscular dystrophies with both paediatric motor signs and late-adult cardiorespiratory manifestations).}
\label{tab:clusters}
\begin{tabular}{lrrrl}
\toprule
\textbf{Archetype} & \textbf{N} & \textbf{Median onset} & \textbf{Median spread} & \textbf{Example diseases} \\
\midrule
Congenital/neonatal & 3{,}573 & 0.5y & 12y & Pierre--Robin, gastroschisis \\
Broad-onset progressive & 4{,}334 & 11y & 78y & Duchenne MD, neurodevelopmental disorders \\
Late-onset & 900 & 54.5y & 20y & ALS, prostate cancer, acute coronary syndrome \\
Early childhood & 128 & 2y & 10y & BPES, O'Donnell--Luria--Rodan syndrome \\
\bottomrule
\end{tabular}
\end{table}

\begin{figure}[h]
    \centering
    \includegraphics[width=0.8\linewidth]{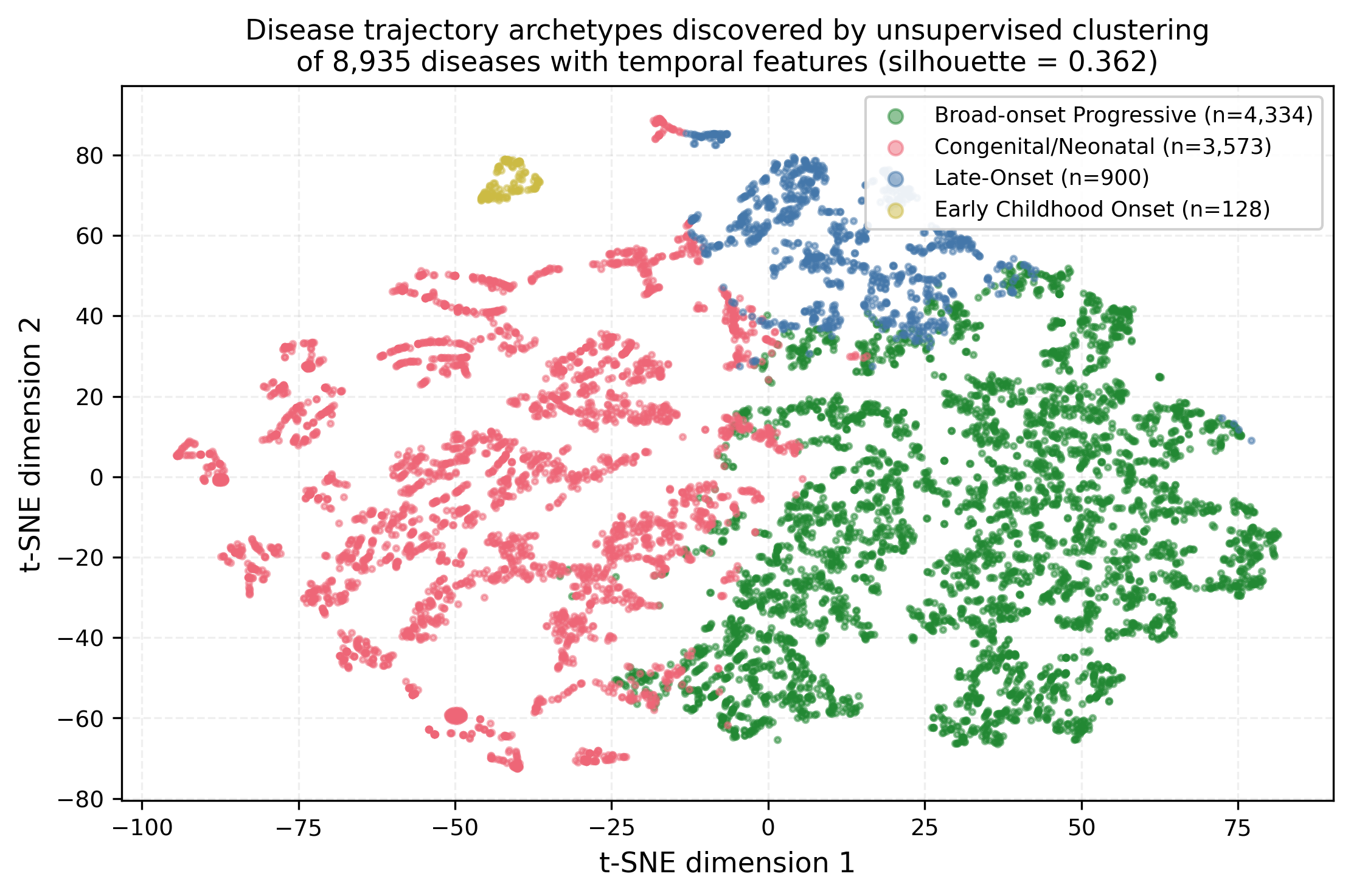}
    \caption{Disease trajectory archetypes discovered by unsupervised clustering of 8{,}935 diseases with temporal features. Four clinically-coherent clusters emerge: congenital/neonatal, broad-onset progressive, early childhood, and late-onset.}
    \label{fig:clusters}
\end{figure}

\paragraph{Sensitivity to clustering choice.} We also ran HDBSCAN on a 4-feature subset (median onset, onset spread, number of stages, total triples) with \texttt{min\_cluster\_size}~$\in\{50, 100, 200, 500, 1000\}$. Small values recover 7--11 clusters but assign 74--79\% of diseases to noise (silhouette of non-noise points 0.41--0.47); large values degenerate to 2 clusters (silhouette 0.71 but structurally uninformative) or return no clusters at all. K-means $k{=}4$ recovers a comparable non-noise silhouette (0.42 on the 4-feature subset) while assigning every disease to an archetype, which is the behaviour required for the coverage-gap and trajectory analyses. The K-means solution is therefore a robustness-preserving reporting choice, not a cherry-picked algorithm.

\subsection{Evidence Decay: Full Age Distribution}
\label{app:decay}

PrimeKG has 4.05M edges and \textbf{zero} evidence dates at the edge level. Each edge is a database snapshot with no way to determine when a fact was established, what evidence supports it, or whether newer findings have refined or contradicted it. A clinician querying PrimeKG for ``DMD $\rightarrow$ corticosteroid treatment'' cannot distinguish a 1995 consensus from a 2024 guideline update. ChronoMedKG, by contrast, has 455{,}519 triples (98.9\% of validated) with PMID-traceable publication dates.

The evidence age distribution reveals that ChronoMedKG integrates knowledge spanning six decades of biomedical research: 27.4\% of triples cite evidence from the last 5 years (2021--2026), reflecting recent discoveries and guideline updates, while 24.1\% cite evidence older than 20 years, capturing established clinical knowledge. The median evidence year is 2015. This temporal provenance enables evaluative questions that static KGs cannot support: ``Is this association based on recent evidence?'' ``Has this treatment recommendation been updated?'' ``What proportion of knowledge about disease X predates the genomic era?''

\begin{figure}[h]
    \centering
    \includegraphics[width=0.7\linewidth]{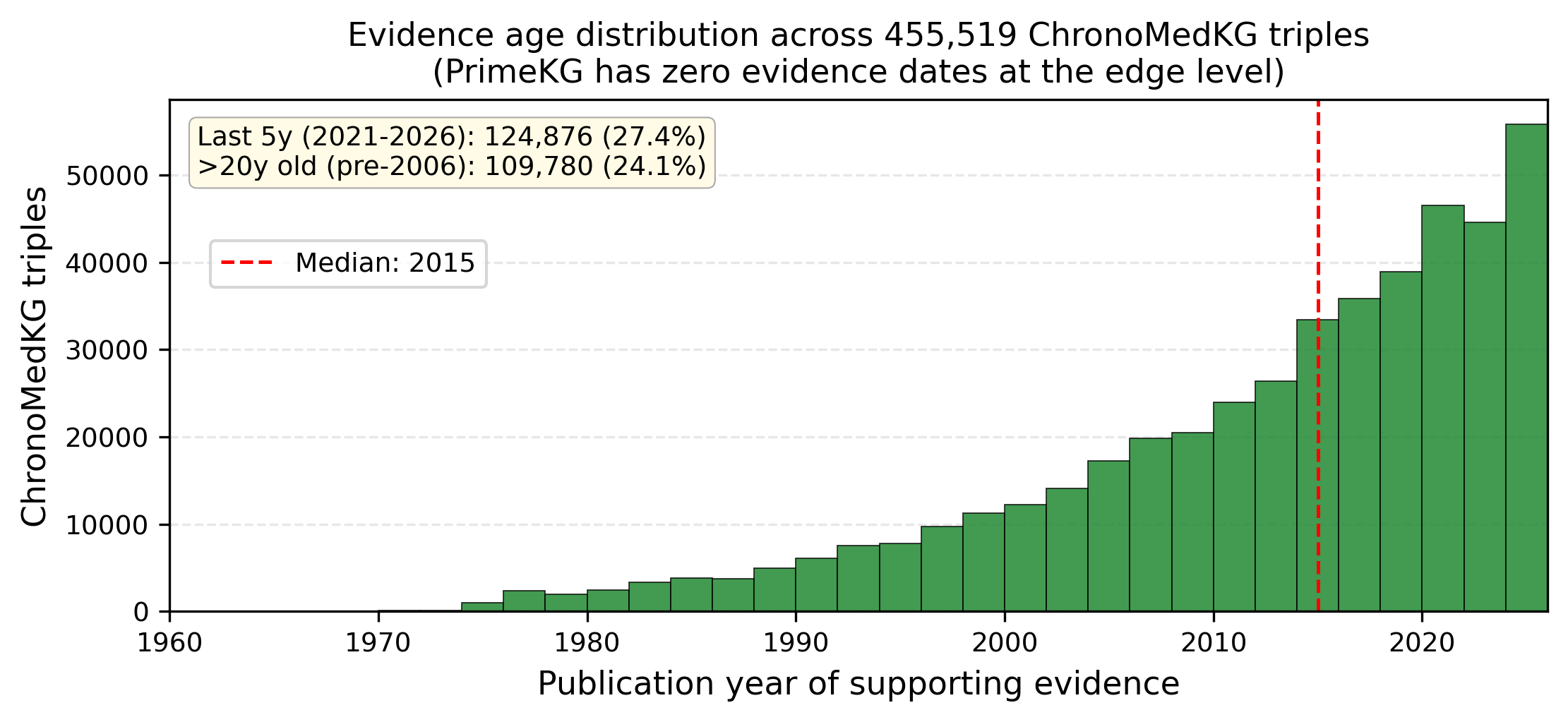}
    \caption{Evidence publication year distribution across 455K ChronoMedKG triples with PMID-traceable dates. Median year: 2015. PrimeKG, by contrast, has zero evidence dates at the edge level, leaving no way to assess recency, supersession, or evidence evolution.}
    \label{fig:decay}
\end{figure}

\section{Extended Ablations}
\label{app:ablations}

\subsection{Link Prediction with Temporal Features: Setup and Full Result}
\label{app:linkpred-full}

To test whether ChronoMedKG's temporal annotations provide useful signal for standard KG evaluation, we trained TransE \citep{transe} via PyKEEN \citep{ali2021pykeen} on disease--phenotype link prediction under four conditions, restricted to the 1{,}302 diseases present in both HPOA and ChronoMedKG for fair comparison: (1) \textbf{HPOA-struct}: plain $(d, \text{has\_phenotype}, p)$ triples; (2) \textbf{HPOA-temporal}: $(d, \text{has\_phenotype\_onset<bin>}, p)$ with HPOA's coarse onset categories (disease-level); (3) \textbf{ChronoMedKG-struct}: plain ChronoMedKG disease--phenotype triples; (4) \textbf{ChronoMedKG-temporal}: ChronoMedKG triples with fine-grained per-phenotype onset bins. Each condition was trained with 3 random seeds (embedding dim 100, 100 epochs, Adam lr=0.01); RotatE \citep{sun2019rotate} and ComplEx were tested but failed on Apple Silicon MPS, see Appendix~\ref{app:linkpred-mps}.

\begin{table}[h]
\centering
\small
\caption{KG link prediction on disease--phenotype edges (TransE, 3 seeds). Temporal features significantly improve MRR in both HPOA and ChronoMedKG, with ChronoMedKG's fine-grained annotations providing larger relative gains than HPOA's coarse bins.}
\label{tab:linkpred}
\begin{tabular}{lrrrr}
\toprule
\textbf{Condition} & \textbf{MRR (mean $\pm$ std)} & \textbf{Hits@1} & \textbf{Hits@10} & \textbf{Gain} \\
\midrule
HPOA-struct & $0.0224 \pm 0.0015$ & 0.004 & $0.0563 \pm 0.006$ & --- \\
HPOA-temporal (coarse) & $0.0389 \pm 0.0001$ & 0.006 & $0.0961 \pm 0.002$ & \textbf{+73.6\%} ($p{=}0.003$) \\
ChronoMedKG-struct & $0.0101 \pm 0.0003$ & 0.001 & $0.0244 \pm 0.002$ & --- \\
ChronoMedKG-temporal (fine) & $0.0192 \pm 0.0015$ & 0.002 & $0.0473 \pm 0.004$ & \textbf{+89.8\%} ($p{=}0.015$) \\
\bottomrule
\end{tabular}
\end{table}

\begin{figure}[h]
\centering
\includegraphics[width=0.7\linewidth]{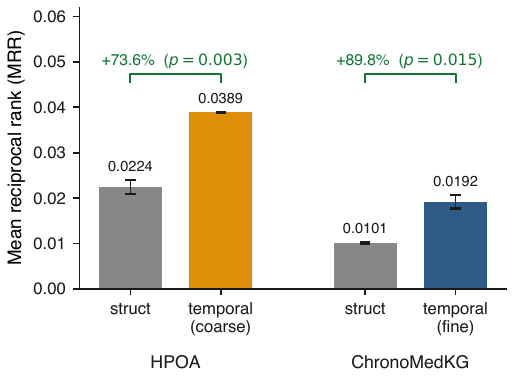}
\caption{Mean reciprocal rank for TransE link prediction on disease--phenotype edges (3 random seeds; error bars are std). Temporal features lift MRR over the matched structural baseline in both HPOA ($+$73.6\%, $p{=}0.003$) and ChronoMedKG ($+$89.8\%, $p{=}0.015$); the larger relative gain in ChronoMedKG sits on a 3$\times$ harder entity space.}
\label{fig:linkpred}
\end{figure}

Temporal features significantly improve MRR in both systems (paired $t$-test across 3 random seeds per condition). The remaining subsections cover scope of claims (\S\ref{app:linkpred-scope}), the MPS-related model coverage (\S\ref{app:linkpred-mps}), and the bin-granularity sensitivity sweep (\S\ref{app:linkpred-sensitivity}).

\subsection{Scope of the Link-Prediction Experiment: Controlled Ablation, Not SOTA Claim}
\label{app:linkpred-scope}

The link-prediction results in Appendix~\ref{app:linkpred-full} are a controlled ablation of temporal features, not a claim that ChronoMedKG outperforms state-of-the-art biomedical KG embeddings. The absolute MRR values (ChronoMedKG-temporal: 0.0192; HPOA-temporal: 0.0389) are low in absolute terms because the experiment deliberately restricts to the intersection of HPOA and ChronoMedKG's disease sets (1{,}302 diseases) and uses the raw LLM-extracted phenotype vocabulary without entity canonicalisation, making ChronoMedKG's entity space roughly $3\times$ larger than HPOA's for the same disease set. A larger entity space makes link prediction harder because the denominator of the rank computation is larger.

\paragraph{Entity-space expansion is a handicap, not an excuse.} The $3\times$ entity space makes ChronoMedKG's TransE task strictly harder than HPOA's on the same disease set. It makes the $+$89.8\% relative gain a harder threshold to clear, not an easier one. ChronoMedKG-temporal must beat ChronoMedKG-struct on a larger, noisier vocabulary than HPOA does.

\paragraph{Absolute MRR is within the biomedical KG range.} Absolute MRR in the 0.01--0.05 range is consistent with other biomedical KG link-prediction studies at similar entity-space sizes (e.g., Hetionet drug-repurposing benchmarks; default PyKEEN TransE training on PrimeKG). A reader unfamiliar with the biomedical KG literature may find 0.0192 surprisingly low; it is not anomalous for this vocabulary scale.

\paragraph{Relative gain replicates across seeds and metrics.} The $+$89.8\% gain is reproducible across 3 random seeds (paired $t$-test, $p{=}0.015$) and holds across MRR, Hits@1, and Hits@10 independently (Table~\ref{tab:linkpred}). A finding that replicates across three metrics and three seeds is robust regardless of absolute magnitude.

Entity canonicalisation via SapBERT/UMLS (Limitation~(i) in main paper Section~7) would compress the ChronoMedKG vocabulary and likely raise absolute MRR substantially; that work is out of scope for v1 and deferred to future work.

\subsection{Consensus Threshold Sensitivity}

The default consensus threshold is 2 (at least 2 distinct LLMs must agree).
We also analyzed results at thresholds 1 (any single model), 2 (default), and 3 (strict).

Note on multi-model agreement: 99.4\% of validated triples have an \texttt{extraction\_models} field containing one model name, while 87.1\% have \texttt{consensus\_confidence}~=~1.00 (full agreement across all models that processed the document). The two fields play different roles: \texttt{extraction\_models} records the model whose surviving extraction was kept as the representative row, and \texttt{consensus\_confidence} records how many independent models agreed on it. The consensus filter uses the latter.

\subsection{Link Prediction: Multiple Models (TransE)}
\label{app:linkpred-mps}

We tested RotatE and ComplEx as additional baselines but both failed due to
PyKEEN/MPS compatibility issues on Apple Silicon (``norm ops not supported
for complex y''). CPU-only training for complex-valued models is slower and
was not pursued at scale. TransE (real-valued) ran successfully on MPS with
the hardware at hand.

\subsection{Bin-Granularity Sensitivity for Link Prediction}
\label{app:linkpred-sensitivity}

To disentangle ``temporal bins help'' from ``fine-grained temporal bins help'', we collapsed ChronoMedKG's eight numeric-derived bins (neonatal, infantile, early childhood, childhood, juvenile, young-adult, adult, late-onset) into five HPOA-style categories by pairwise merge (antenatal-infantile, childhood, juvenile, adult, late-onset) and retrained under identical conditions.

\begin{table}[h]
\centering
\small
\caption{Bin-granularity sensitivity on ChronoMedKG disease--phenotype link prediction (3 seeds per cell). \emph{Coarse} = 5 HPOA-style bins; \emph{fine} = 8 numeric-derived bins. $\Delta$ = relative MRR gain over the no-temporal \texttt{ta\_struct} baseline (TransE 0.0105$\pm$0.0005) for TransE rows; paired $t$-test rightmost column compares fine vs coarse within-scorer.}
\label{tab:linkpred-sensitivity-supp}
\setlength{\tabcolsep}{4pt}
\begin{tabular}{llrrrl}
\toprule
\textbf{Scorer} & \textbf{Bins} & \textbf{MRR} & \textbf{Hits@10} & \textbf{$\Delta$ struct} & \textbf{fine vs coarse} \\
\midrule
TransE   & coarse (5) & $0.0177 \pm 0.0003$ & $0.0445 \pm 0.0006$ & $+$68.4\% & --- \\
TransE   & fine (8)   & $0.0182 \pm 0.0007$ & $0.0440 \pm 0.0033$ & $+$72.9\% & $+$2.7\%, $p{=}0.42$ \\
\midrule
DistMult & coarse (5) & $0.0133 \pm 0.0011$ & $0.0258 \pm 0.0024$ & ---       & --- \\
DistMult & fine (8)   & $0.0157 \pm 0.0022$ & $0.0327 \pm 0.0043$ & ---       & $+$18.2\% \\
\bottomrule
\end{tabular}
\end{table}

The sensitivity result is a negative finding for \emph{fine} granularity specifically but a strongly positive one for temporal metadata overall. On TransE, coarse 5-bin ChronoMedKG and fine 8-bin ChronoMedKG are indistinguishable ($+$2.7\% mean, paired $t{=}1.02$, $p{=}0.42$); the $+$72.9\% lift over \texttt{ta\_struct} is already captured by five bins. On DistMult, fine 8-bin shows a $+$18.2\% advantage over 5-bin, suggesting that the utility of finer bins depends on scorer expressivity. The load-bearing property is that ChronoMedKG encodes per-phenotype onset windows at all. Entity-name canonicalisation (Limitation~(i) in main text Section~7) is the more pressing remaining gap: with a 3$\times$ smaller entity space, fine-grained bins may recover a larger signal than we observed here.

\newpage

\section{Data, Code, and Reproducibility}
\label{app:availability}

\subsection{Dataset Availability}

\begin{itemize}
    \item \textbf{ChronoMedKG dataset} (460{,}497 validated triples derived from 13{,}431 PrimeKG diseases), organised as three tiers matching the construction pipeline:
        \begin{itemize}
            \item \textbf{Concept DOI} (always latest): \url{https://doi.org/10.5281/zenodo.19697542}
            \item \textbf{Version DOI} (v0.0.1 reviewed here): \url{https://doi.org/10.5281/zenodo.19697543}
            \item Gold tier: \texttt{validated\_triples.jsonl} (502 MB, 460{,}497 post-QC triples)
            \item Silver tier: \texttt{consensus\_triples.jsonl.gz} (30 MB, 443{,}114 pre-QC multi-LLM consensus rows)
            \item Bronze tier: \texttt{raw\_triples.jsonl.gz} (644 MB, 13{,}001{,}651 raw LLM extractions across four models)
            \item Silver and Bronze rows carry a \texttt{disease\_mondo} field so each row is self-locating
            \item Hugging Face mirror with \texttt{datasets.load\_dataset()} API: deferred to camera-ready; during review, reviewers download directly from Zenodo
            \item License: CC BY 4.0
            \item Format: JSONL (one triple per line); Silver/Bronze tiers gzipped
            \item Croissant 1.0 metadata (8 RAI fields) shipped alongside the data for mlcommons tooling
            \item README includes a \emph{Known Discrepancies} section documenting audit findings, all under 1\% of their respective denominators: 4{,}331 validated rows in the flat file that do not appear in the per-disease tree (0.94\% of 460{,}497), 54 diseases with validated rows that bypassed the consensus intermediate (0.50\% of 10{,}852), 44{,}036-row drift between pre- and post-merge raw counts (0.34\% of 13M)
        \end{itemize}

    \item \textbf{ChronoTQA benchmark} (3{,}341 questions across 9 task types):
        \begin{itemize}
            \item Co-released at the same Zenodo deposit (\texttt{tqa\_benchmark.json}, 3 MB)
            \item JSON format with answers and gold-standard metadata (Tier 1 = external source, Tier 2 = ChronoMedKG-derived with PMID trace)
        \end{itemize}

    \item \textbf{PMC clinical cases} (31 curated cases):
        \begin{itemize}
            \item Shipped in the Zenodo deposit as \texttt{pmc\_clinical\_cases.json} (63 KB)
            \item Each case is a JSON record linked to its PMC URL (all open-access)
        \end{itemize}
\end{itemize}

\subsection{Code Availability}

\begin{itemize}
    \item \textbf{Pipeline code:} \url{https://gitlab.sdu.dk/screen4care/chronomedkg}
    \item \textbf{License:} MIT
    \item \textbf{Language:} Python 3.12
    \item \textbf{Key dependencies:} OpenAI, Anthropic, Google-GenerativeAI, DeepSeek SDKs; PyKEEN 1.11; scikit-learn; rapidfuzz; pandas
    \item \textbf{Reproduce link prediction:} \texttt{.venv-sapbert/bin/python scripts/experiment\_link\_prediction\_v3.py}
    \item \textbf{Reproduce all experiments:} see \texttt{scripts/experiment\_*.py}
\end{itemize}

\subsection{Compute Requirements}

\begin{itemize}
    \item \textbf{Pipeline extraction} (one-time): $\sim$\$2{,}400 LLM API costs, $\sim$100 CPU-hours
    \item \textbf{Link prediction experiment}: 3 minutes on Apple M4 MPS, no GPU required
    \item \textbf{Trajectory clustering}: 2 minutes on CPU (sklearn K-Means + TSNE)
    \item \textbf{Evidence decay analysis}: 5 minutes (single-pass over JSONL files)
\end{itemize}

\subsection{Reproducibility Statement}

All experiments use fixed random seeds (42, 7, 123 for link prediction; 42 for clustering). Exact package versions are in \texttt{requirements.txt}. The decision log at \texttt{docs/decision\_log.md} tracks all methodological changes during development.

\section{Ethics Statement}
\label{app:ethics}

\textbf{Data provenance.} ChronoMedKG is constructed entirely from (i)~public biomedical literature (PubMed abstracts and PMC Open Access full-text, via NCBI E-utilities with an API key), (ii)~publicly-licensed ontologies and knowledge bases (PrimeKG, Orphadata, HPOA, GA4GH Phenopackets), and (iii)~LLM-generated extractions from those sources via commercial APIs (OpenAI, Anthropic, Google, DeepSeek). No patient-level records, electronic health records, images, or identifiers are used at any stage of the pipeline or included in the released resource.

\textbf{Clinician evaluation.} The three-clinician panel results we cite (Appendix~\ref{app:clinician}) were collected in a prior study (HEG-TKG~\citep{ahmed2026hegtkg}) on rare neuromuscular diseases. Ratings were gathered from three physicians (one senior neurologist, one physician-scientist, one medical-informatics specialist) who consented to their anonymised Likert ratings being used for research purposes, including the comparative analysis reported here. The rating task assessed system-generated literature summaries, not patient data; no individual patient was the subject of any rating. Institutional ethics review was not required for the clinician-rating component under the rules of the collaborating institutions because the activity did not involve human-subjects research beyond expert opinion on synthetic outputs. Per-rater identifiers are anonymised to C1/C2/MI and only aggregate statistics (means, effect sizes) are reported.

\textbf{Clinical case studies.} The 31 diagnostic-odyssey cases cited in the main text (and listed in full in Appendix~\ref{app:pmc-cases}) are all taken from PubMed Central open-access case reports. Each case is referenced only by its PMC identifier; no re-identification is performed, no patient-contact information is collected, and the illustrative description in the paper consists of disease-course summaries already publicly disclosed in the source report. No new patient recruitment occurred.

\textbf{LLM use and disclosure.} LLMs are used in two distinct roles in this work: as components inside the construction pipeline, and as writing assistants during manuscript preparation. Both are disclosed below.

\emph{Pipeline LLMs (extraction and judging).} Extractions are produced by commercial LLM APIs: DeepSeek V3 and an OpenAI primary (GPT-4.1-nano or GPT-4o-mini), with Claude 3 Haiku as a conditional tiebreaker; the three-LLM novelty-judge panel uses DeepSeek V3, GPT-4o-mini, and Claude Haiku 4.5. We record the model identifier, prompt, and timestamp for every extracted triple; the representative-model field documented in these supplementary methods is paired with \texttt{consensus\_confidence}, which records full multi-LLM agreement. LLM output is treated as a literature summary, not an authoritative clinical reference; 7.3\% of outputs are demonstrably wrong (main text Section~4.3), and released triples carry per-source PMIDs so users can verify every claim.

\emph{Writing-assistant LLMs (manuscript preparation).} Claude (Anthropic) assisted the authors with prose drafting and revision against author-prepared outlines and analyses, structural restructuring of sections, and cross-checking references against their sources. Grammarly and Overleaf were used for copyediting. AI-assisted prose was reviewed by the authors against the underlying experimental results before inclusion. No LLM was used to generate numerical results, to perform statistical analysis, or to make experimental-design decisions. All scientific content, experimental design, data collection, analysis, and interpretation were performed by the authors, who take full responsibility for the manuscript content.

\emph{Image-generation AI (figures).} Figure~1 in the main paper (the four-agent pipeline schematic) was produced with assistance from a diagram-generation AI tool; the resulting schematic was reviewed and refined by the authors for technical accuracy before inclusion. Any schematic styling or layout assistance in other figures has likewise been reviewed by the authors against the underlying experimental setup. No data-bearing element in any figure was generated or modified by AI: every quantitative figure in this paper (Figure~3 in the main text and all data plots in the supplementary) is rendered directly from the released benchmark data via Python scripts shipped in the code repository, with no AI image-generation tool in the rendering pipeline.

\textbf{Risks and safeguards.} ChronoMedKG is a literature knowledge graph, not a clinical-decision tool. Direct downstream use in patient-facing systems without clinician oversight is explicitly not supported. The resource has not been regulated as a medical device and must not be used as one. Downstream users building clinical tools that incorporate ChronoMedKG are responsible for their own regulatory clearance, clinician review, and patient-safety evaluation. The release includes the main-text error-rate table and the novel-coverage verification protocol so users can bound expected-error budgets in their own deployments.

\textbf{Dual use.} Biomedical knowledge graphs could in principle be used to target patients in harmful ways, such as risk-based insurance discrimination by age-of-onset. ChronoMedKG contains only population-level onset ranges drawn from published literature, information that is already public, and adds no private or patient-linked information. We do not identify any dual-use risk beyond those already present in the underlying literature.

\section{Reviewer Access and Release Artefacts}
\label{app:hf}

\paragraph{Data access.} ChronoMedKG v0.0.1 is released open-access under CC~BY~4.0 on Zenodo. The \textbf{concept DOI} \url{https://doi.org/10.5281/zenodo.19697542} always resolves to the latest version; the \textbf{version DOI} \url{https://doi.org/10.5281/zenodo.19697543} resolves specifically to v0.0.1 (the bytes reviewed for this submission). A Hugging Face mirror exposing the \texttt{datasets} \texttt{load\_dataset()} API will be made available at camera-ready; during review, reviewers should download directly from Zenodo. Code (the four-agent pipeline and experiment scripts) is released separately under the MIT License.

\paragraph{Release artefacts.} The Zenodo deposit contains nine files, organised as three tiers matching the construction pipeline: (1)~\texttt{validated\_triples.jsonl} (502~MB, 460{,}497 Gold-tier post-QC triples); (2)~\texttt{consensus\_triples.jsonl.gz} (30~MB, 443{,}114 Silver-tier pre-QC consensus rows); (3)~\texttt{raw\_triples.jsonl.gz} (644~MB, 13{,}001{,}651 Bronze-tier raw LLM extractions across four models); (4)~\texttt{tqa\_benchmark.json} (3~MB, 3{,}341 ChronoTQA questions); (5)~\texttt{pmc\_clinical\_cases.json} (63~KB, 31 diagnostic-odyssey cases); (6)~\texttt{croissant.json} (Croissant~1.0 metadata with the eight RAI fields: \texttt{dataCollection}, \texttt{dataPreprocessingImputation}, \texttt{dataUseCases}, \texttt{dataLimitations}, \texttt{dataSocialImpact}, \texttt{annotationsPerItem}, \texttt{personalSensitiveInformation}, \texttt{dataReleaseTimeline}); (7)~\texttt{README.md} (data card with full schema, tier descriptions, and a Known Discrepancies section); (8)~\texttt{LICENSE-DATA} (CC~BY~4.0); (9)~\texttt{NOTICE} (attribution to upstream sources including PrimeKG, PubMed, PMC, MONDO, OMIM, HPO).

\paragraph{Code and environment.} Code is released under the MIT License (see \texttt{LICENSE-CODE}). A pinned \texttt{requirements.txt} fixes the versions of 15 load-bearing Python packages used to produce this release (\texttt{openai==2.9.0}, \texttt{anthropic==0.75.0}, \texttt{rapidfuzz==3.14.3}, \texttt{torch==2.7.0}, etc.); a companion \texttt{.venv-sapbert} environment is reused unchanged from the HEG-TKG project for SapBERT-based entity normalisation. Random seeds are fixed where applicable (link prediction: $\{42,7,123\}$; novelty sampling: $42$).

\end{document}